\pdfoutput=1
\documentclass[11pt]{article}

\usepackage[final]{ACL2023}

\usepackage{float}
\usepackage{times}
\usepackage{latexsym}
\usepackage{epigraph} 
\usepackage[most]{tcolorbox}
\usepackage{enumitem}
\usepackage[rgb]{xcolor}
\usepackage{booktabs}
\usepackage{multirow}
\usepackage{pifont}
\usepackage{xspace}
\usepackage{tabularx}
\usepackage{wrapfig}
\usepackage[normalem]{ulem}
\useunder{\uline}{\ul}{}

\definecolor{customblue}{rgb}{0.168, 0.364, 0.557}
\definecolor{customgreen}{rgb}{0.0, 0.5, 0.0}

\usepackage[T1]{fontenc}
\usepackage[utf8]{inputenc}

\usepackage{microtype}

\usepackage{inconsolata}

\usepackage{graphicx}
\usepackage{xspace}
\usepackage{amsmath}
\usepackage{amssymb}
\usepackage{pifont}
\usepackage{multirow}
\usepackage{threeparttable}
\usepackage{tabularx}
\usepackage{booktabs}
\usepackage{makecell}

\usepackage{graphicx}
\usepackage[table,xcdraw]{xcolor}
\usepackage{colortbl}

\newcommand{\method}{\textsc{Q-Daps}\xspace}

\newcolumntype{Y}{>{\centering\arraybackslash}X}

\title{Question Difficulty Estimation for Large Language Models via Answer Plausibility Scoring}
\usepackage{orcidlink}

\newcommand{\JMorcid}{\orcidlink{0000-0003-4850-9239}}
\newcommand{\BPorcid}{\orcidlink{0009-0005-3578-2393}}
\newcommand{\AJorcid}{\orcidlink{0000-0001-7235-0665}}

\author{Jamshid Mozafari\thanks{\, Corresponding Author.}~~\JMorcid, Bhawna Piryani \BPorcid, Adam Jatowt \AJorcid \\
  University of Innsbruck, Innsbruck, Austria \\
  \texttt{\{jamshid.mozafari, bhawna.piryani, adam.jatowt\}@uibk.ac.at} \\
}

\begin{document}
\maketitle
\begin{abstract}
Estimating question difficulty is a critical component in evaluating and improving large language models (LLMs) for question answering (QA). Existing approaches often rely on readability formulas, retrieval-based signals, or popularity statistics, which may not fully capture the reasoning challenges posed to modern LLMs. In this paper, we introduce \method (\textbf{Q}uestion \textbf{D}ifficulty based on  \textbf{A}nswer \textbf{P}lausibility \textbf{S}cores) method, a novel approach that estimates question difficulty by computing the entropy of plausibility scores over candidate answers.
We systematically evaluate \method across four prominent QA datasets—TriviaQA, NQ, MuSiQue, and QASC—demonstrating that it consistently outperforms baselines. Moreover, \method shows strong robustness across hyperparameter variations and question types. Extensive ablation studies further show that \method remains robust across different plausibility estimation paradigms, model sizes, and realistic settings. Human evaluations further confirm strong alignment between \method’s difficulty estimates and human judgments of question difficulty. Overall, \method provides an interpretable, scalable, and bias-resilient approach to question difficulty estimation in modern QA systems.

\end{abstract}

\noindent 
\begin{wrapfigure}{l}{0.03\textwidth} \centering 
\hypertarget{github-link}{} \href{https://github.com/DataScienceUIBK/Q-DAPS}{\includegraphics[width=0.05\textwidth]{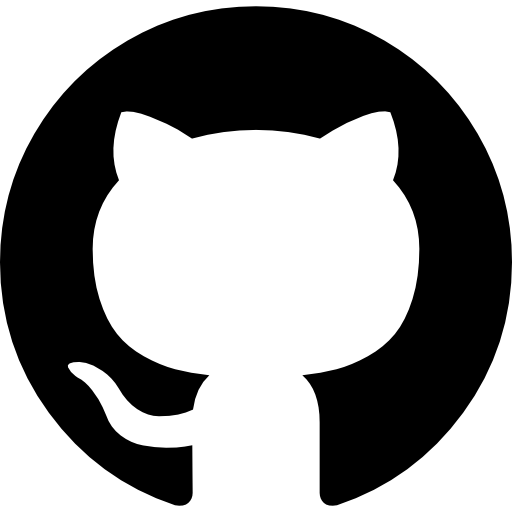}} \vspace{-3em} 
\end{wrapfigure}

\noindent {\fontsize{10}{10}\selectfont\url{https://github.com/DataScienceUIBK/Q-DAPS}}

%%% INTRODUCTION %%%

\section{Introduction}\label{s:introduction}

Questions are a fundamental means by which users express their information needs in Information Retrieval (IR) and Natural Language Processing (NLP) systems. They span a wide range of types—factoid, definition, and yes/no~\cite{pandya2021questionansweringsurveydirections}—and can also be classified by difficulty~\cite{10.1145/3556538}, such as easy, medium, or hard~\cite{raina2024questiondifficultyrankingmultiplechoice}, or as simple vs. complex~\cite{gabburo-etal-2024-measuring}.

Question difficulty reflects how challenging it is to answer a given question. Prior work has explored difficulty estimation using features such as readability~\cite{naous-etal-2024-readme}, retrieval-based signals~\cite{gabburo-etal-2024-measuring}, and metrics derived from large language models (LLMs)~\cite{2024.EDM-posters.90}. However, most of these studies are primarily focused on estimating difficulty from the perspective of human readers or retrieval systems.

\begin{figure}
	\centering
	\includegraphics[width=\columnwidth]{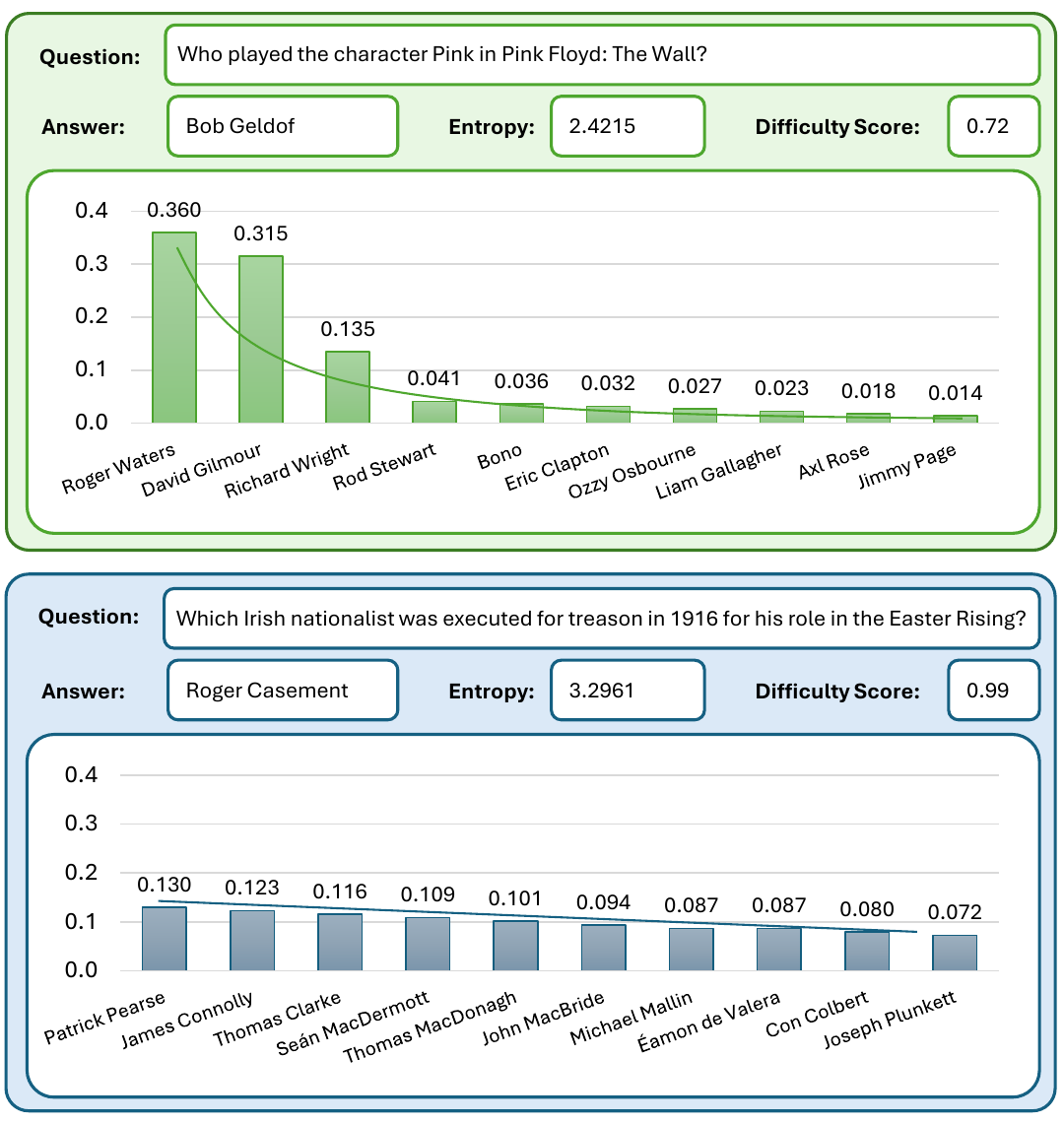}
	\caption{Two examples from the NQ (\textcolor{green}{green}) and TriviaQA (\textcolor{blue}{blue}) datasets are presented, each showing the correct answer, 10 candidate answers (selected from 20 generated candidates), their normalized plausibility scores, and the computed difficulty score. The \textcolor{green}{green} colored example illustrates low entropy (an easier question), while the \textcolor{blue}{blue} colored example demonstrates high entropy (a harder question).}
	\label{fig:teaser}
\end{figure}
\begin{figure*}
	\centering
	\includegraphics[width=\linewidth]{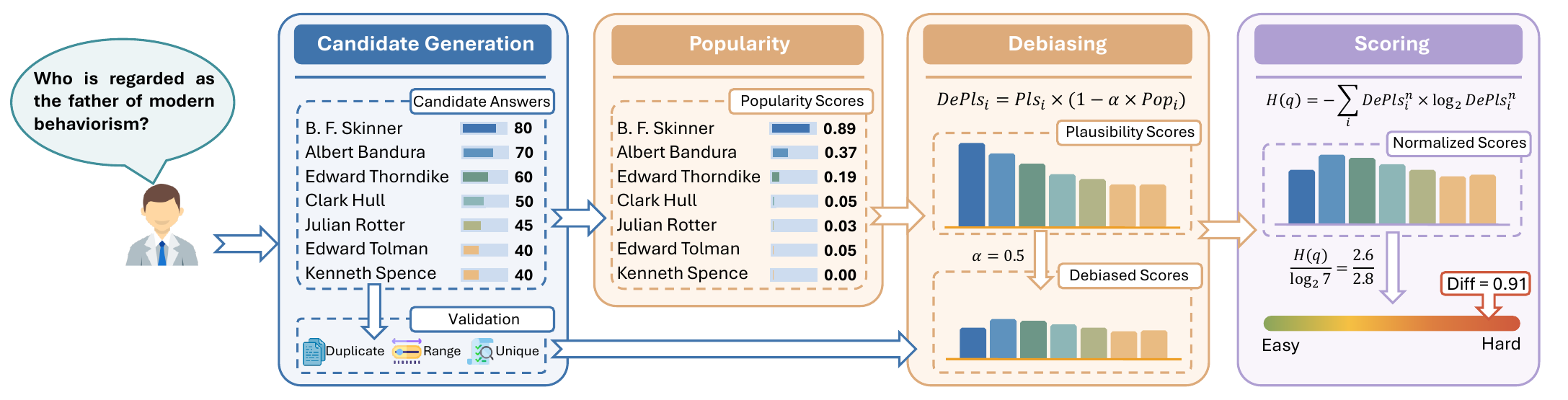}
	\caption{The \method method comprises three stages: \textbf{\textcolor{blue}{Candidate Generation}}, which produces candidate answers and their plausibility scores; \textbf{\textcolor{orange}{Popularity Debiasing}}, which adjusts these plausibility scores based on candidate popularity; and \textbf{\textcolor{violet}{Scoring}}, which computes the final difficulty score for the given question.
}
	\label{fig:method}
\end{figure*}

Our work differs in two key ways: (1) it provides an LLM-oriented difficulty measure grounded in how convincing incorrect answers appear to the LLM, rather than in surface-level text properties designed for humans or retrieval statistics tailored to IR systems, and (2) it directly links difficulty estimation to a critical application—hallucination risk detection—where harder questions are more likely to elicit fabricated or incorrect answers. This operational perspective motivates our method as a practical tool in high-stakes deployments: for model selection (e.g., choosing a stronger LLM if most domain questions are difficult), question routing (e.g., sending high-difficulty questions to a human reviewer in a company knowledge base), and safeguard triggering (e.g., requiring citations or user confirmation for exam questions).\footnote{In Appendix~\ref{s:related_works}, we provide a more detailed discussion of related works and compare our method with other works.}

In this paper, we introduce \textit{\textbf{Q}uestion \textbf{D}ifficulty based on \textbf{A}nswer \textbf{P}lausibility \textbf{S}cores} (\method), a new method for estimating question difficulty tailored to LLMs. The core idea is to leverage candidate answers\footnote{Candidate answers refer to plausible but incorrect answers for a given question.} and their associated plausibility scores~\cite{10.1145/3726302.3730299}, which reflect how convincing each incorrect answer is. 
For example, for the question \textit{What is the capital of China?}, the correct answer is \textit{Beijing}, but two example candidates such as \textit{Shanghai} and \textit{Shenzhen} might also seem plausible, even though they are incorrect. However, \textit{Shanghai} may appear more convincing due to its global prominence as China’s largest city, and should therefore receive a higher plausibility score than \textit{Shenzhen}.

We hypothesize that the entropy~\cite{https://doi.org/10.1002/j.1538-7305.1948.tb01338.x} of normalized plausibility scores provides an effective signal for estimating question difficulty. High entropy suggests that many candidate answers are similarly plausible, indicating a harder question. In contrast, low entropy reflects a skewed distribution, where only a few candidates are highly plausible—indicating an easier question. Figure~\ref{fig:teaser} shows two examples from Natural Questions~\cite{kwiatkowski-etal-2019-natural} and TriviaQA~\cite{joshi-etal-2017-triviaqa} datasets, with their normalized plausibility scores of candidate answers and the estimated difficulty scores.

An additional challenge in this setting is popularity bias~\cite{Klimashevskaia2024}, which is well-known in recommender systems and has also been observed in LLM-generated outputs~\cite{ni2025knowledgepopularityinfluencesenhances, 10.1145/3726302.3730222, mallen-etal-2023-trust}. Yet, its impact on candidate answers and their plausibility scores for difficulty estimation remains unexplored. We address this gap by explicitly modeling and reducing popularity bias using Wikipedia page view statistics, to improve accuracy of \method method.

As shown in Figure~\ref{fig:method}, the \method proceeds in three stages. First, we generate candidate answers and assign plausibility scores\footnote{Standard distractor generation for multiple-choice QA~\cite{alhazmi-etal-2024-distractor} is unsuitable here, since it labels distractors as correct/incorrect without considering plausibility
~\cite{raina-etal-2023-assessing}.} and validate them. Second, we measure their popularity using Wikipedia page view counts and adjust the plausibility scores to mitigate popularity bias. Finally, we compute the entropy of the debiased scores and normalize it to $[0,1]$, producing an interpretable difficulty estimate.
The estimate is interpretable because, in addition to the difficulty score, we return the candidate answers together with their plausibility scores, allowing users to understand how the difficulty was derived.

We evaluate \method on a diverse set of datasets, covering simple question answering benchmarks such as TriviaQA~\cite{joshi-etal-2017-triviaqa} and Natural Questions~\cite{kwiatkowski-etal-2019-natural}, as well as more complex reasoning benchmarks including MuSiQue~\cite{trivedi-etal-2022-musique} and QASC~\cite{Khot_Clark_Guerquin_Jansen_Sabharwal_2020}. We explore multiple scoring paradigms---pointwise, pairwise, and listwise---and compare \method against a broad range of baseline methods. We also perform extensive ablation studies to examine the effects of different LLMs used for candidate answer generation and plausibility scoring, the role of the \textit{Popularity Debiasing} component, and the impact of providing the \textit{gold answer} during inference. The results demonstrate that the central hypothesis of \method is robust: it performs effectively even without access to the gold answer, without popularity debiasing, and across different underlying LLMs. We further support our conclusions with two human evaluation studies reported in Section~\ref{apx:human_evaluation}. Finally, we present an error analysis (Appendix~\ref{apx:error_analysis}) and a frequently asked questions (FAQ) section (Appendix~\ref{apx:faq}) to address common reader concerns.

\begin{figure}[t]
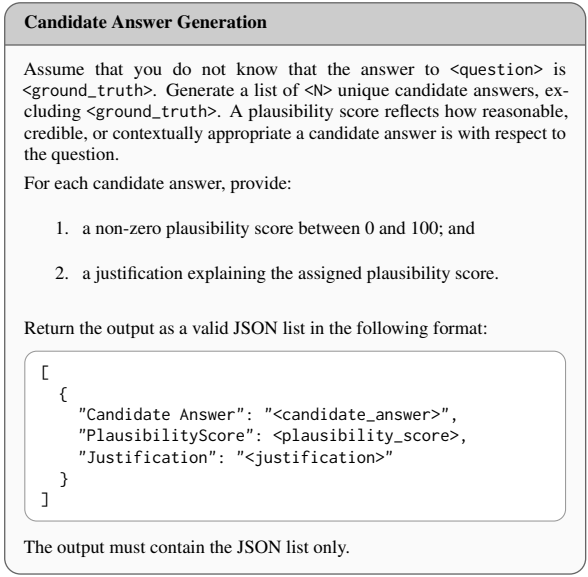

\centering
\begin{tcolorbox}[
    enhanced,
    width=\columnwidth,
    colback=black!2,
    colframe=black!60,
    boxrule=0.5pt,
    arc=2mm,
    left=1.5mm,
    right=1.5mm,
    top=1.2mm,
    bottom=1.2mm,
    title=\textbf{\raisebox{0pt}[2ex][1ex]{Candidate Answer Generation}},
    fonttitle=\bfseries\scriptsize,
    colbacktitle=black!20,
    coltitle=black,
    boxed title style={
        colframe=black!60,
        colback=black!10,
        boxrule=0.5pt,
        arc=2mm,
        valign=center
    }
]
\scriptsize

Assume that you do not know that the answer to \texttt{<question>} is 
\texttt{<ground\_truth>}. Generate a list of \texttt{<N>} unique candidate 
answers, excluding \texttt{<ground\_truth>}.
A plausibility score reflects how reasonable, credible, or contextually 
appropriate a candidate answer is with respect to the question.

\vspace{0.4em}

For each candidate answer, provide:
\begin{enumerate}
    \item a non-zero plausibility score between 0 and 100; and
    \item a justification explaining the assigned plausibility score.
\end{enumerate}

\vspace{0.5em}

Return the output as a valid JSON list in the following format:
\begin{tcolorbox}[
    colback=white,
    colframe=black!40,
    boxrule=0.4pt,
    arc=1.5mm,
    left=1mm,
    right=1mm,
    top=0.8mm,
    bottom=0.8mm
]
\ttfamily\scriptsize
\begin{verbatim}
[
  {
    "Candidate Answer": "<candidate_answer>",
    "PlausibilityScore": <plausibility_score>,
    "Justification": "<justification>"
  }
]
\end{verbatim}
\end{tcolorbox}

\vspace{0.3em}

The output must contain the JSON list only.

\end{tcolorbox}

\caption{Prompt used for listwise candidate answer generation. 
\texttt{<question>} denotes the input question, \texttt{<ground\_truth>} the 
correct answer, and \texttt{<N>} the number of candidates. Each 
\texttt{<candidate\_answer>} is associated with a plausibility score 
(\texttt{<plausibility\_score>}) and a justification 
(\texttt{<justification>}).}
\label{fig:listwise_prompt}
\end{figure}

\paragraph{Our contributions are as follows:}
\begin{enumerate}
\item We propose \method, a novel LLM-oriented question difficulty metric based on the entropy of answer plausibility scores.
\item We show that popularity bias affects candidate answers and introduce a lightweight debiasing strategy that improves difficulty estimation.
\item We validate \method through extensive experiments across multiple datasets, scoring settings, models, and human evaluations.
\end{enumerate}

%%% METHOD %%%

\section{\method Method}\label{s:method}

\subsection{Candidate Generation}\label{ss:candidate_generation}

In the first stage, we prompt an LLM\footnote{We use LLaMA~3.3~\cite{grattafiori2024llama3herdmodels} as the default generation core; however, the method is not heavily dependent on the choice of LLM, as demonstrated in the ablation study in the Section~\ref{ss:ablation_study}.} to generate $N$ candidate answers\footnote{We selected 20 as an upper limit because, in practice, LLMs are rarely able to generate more than 20 reasonable candidates for most questions.} along with their plausibility scores for a question. 
To generate candidate answers and their plausibility scores~\cite{10.1145/3726302.3730299}, we use the prompt shown in Figure~\ref{fig:listwise_prompt}. 
We pass both the question and its gold answer to the \textit{Candidate Generation} component, which is responsible for candidate generation. In this component, we also request justifications for the plausibility scores, as providing explanations has been shown to improve the reliability of generated outputs~\cite{huang-etal-2023-large}.

We pass the generated candidate answers and their plausibility scores to \textit{Validation} component, which ensures their quality through the following steps: (1) ensuring there are no duplicate candidate answers, (2) verifying that plausibility scores fall within the valid range of $0$ to $100$, and (3) confirming that exactly $N$ unique candidate answers are produced. Duplicates are identified using semantic similarity using the BEM method~\cite{bulian-etal-2022-tomayto}. If any validation check fails, we increment the LLM’s temperature by $0.1$ and prompt it to regenerate the list. This process is repeated iteratively until all validation criteria are satisfied.

\subsection{Popularity Debiasing}\label{ss:popularity_debiasing}
In this stage, we extract monthly Wikipedia page view counts for the generated candidate answers. These candidate answers are passed to the \textit{Popularity} component, which computes their popularity based on the number of views of the corresponding Wikipedia page\footnote{If a candidate answer does not have an associated Wikipedia page, its popularity is set to zero.}. 
We use page view data spanning from January 1, 2015, to December 31, 2024, providing a fixed and consistent basis for computing popularity values.

Given the high variability in Wikipedia page view counts, we normalize popularity scores to the $[0, 1]$ range. Outliers are removed using the interquartile range (IQR) method, following the approach of~\citet{mozafari-triviahg}. To efficiently scale this computation, we leverage the HintEval~\cite{mozafari2025hinteval} toolkit, which enables parallel processing and significantly improves the runtime performance of \method.

After computing the popularity scores, each candidate answer—along with its plausibility score and popularity value—is passed to the \textit{Debiasing} component. This module adjusts the plausibility score of every candidate answer based on its popularity, following Equation~\ref{eq:debiasing}:

{\footnotesize
\begin{align}
    \label{eq:debiasing}
    DePls_i &= Pls_i - \alpha \times Pop_i \times Pls_i \\
            &= Pls_i \times (1 - \alpha \times Pop_i) \notag
\end{align}
}

where $Pls_i$ denotes the plausibility score of the $i^{th}$ candidate answer, and $Pop_i$ is its popularity. The hyperparameter $\alpha$ controls the strength of the debiasing adjustment. We multiply $Pop_i$ by $Pls_i$ to ensure that the debiasing effect is relative to the candidate's plausibility—since popularity is a fixed attribute, while plausibility can vary across questions. This computation is also parallelized to further increase the efficiency of \method.

\subsection{Score}\label{ss:score}

In the final stage, we compute the entropy of the debiased plausibility scores (\(DePls\)) for all candidate answers.
First, we normalize the scores to form a valid probability distribution:

\begin{equation}
    \footnotesize
    \label{eq:pop_distribution}
    DePls_i^{\text{norm}} = \frac{DePls_i}{\sum_i DePls_i}
\end{equation}

to later compute the entropy: %using Equation~\ref{eq:entropy}:  

\begin{equation}
    \small
    \label{eq:entropy}
    H(q) = -\sum_i DePls_i^{\text{norm}} \times \log_2 DePls_i^{\text{norm}}
\end{equation}

where \(DePls_i^{\text{norm}}\) denotes the normalized debiased plausibility score of the \(i^{\text{th}}\) candidate answer. Finally, the computed entropy is passed to the \textit{Normalization} component, which scales it to the range $[0, 1]$:\looseness=-1
%according to Equation~\ref{eq:normalization}:  

\begin{equation}
    \small
    \label{eq:normalization}
    Diff_q = \frac{H(q)}{\log_2 N}
\end{equation}

where \(N\) is the total number of candidate answers. For examples of the approach, we refer readers to Appendix~\ref{apx:case_study}.
%, which provides a detailed example of the method.

%%% EXPERIMENTAL SETUP %%%

\section{Experimental Setup}\label{s:experimental_seup}

\subsection{Datasets}\label{ss:datasets}
We use TriviaQA~\cite{joshi-etal-2017-triviaqa} and Natural Questions (NQ) \cite{kwiatkowski-etal-2019-natural} datasets for simple questions, as well as MuSiQue \cite{trivedi-etal-2022-musique} and QASC~\cite{Khot_Clark_Guerquin_Jansen_Sabharwal_2020} for complex questions. From each dataset, we sampled 2,000 questions using stratified sampling~\cite{ARNAB2017213} based on question types, which we identified following the method of~\citet{tayyar-madabushi-lee-2016-high} to have a fair and balanced contribution from each dataset. We use only the questions and their corresponding gold answers from these datasets. Additional details are provided in Appendix~\ref{apx:dataset_details}.\looseness=-1

\begin{table*}
\centering
\resizebox{0.7\textwidth}{!}{%
\begin{tabular}{l@{\hspace{20mm}}|cc|cc|cc|cc}
\toprule
\textbf{Scenario} & 
\multicolumn{2}{c}{\textbf{MuSiQue}} & 
\multicolumn{2}{c}{\textbf{QASC}} & 
\multicolumn{2}{c}{\textbf{NQ}} & 
\multicolumn{2}{c}{\textbf{TriviaQA}} \\
 &  $d$ & $\rho$ & $d$ & $\rho$ & $d$ & $\rho$ & $d$ & $\rho$ \\
\midrule
Pointwise
& 0.0023 & -0.1272 & 0.6214 & -0.4848 & 1.098 & -0.9 & 0.5557 & -0.309 \\
Pairwise
& 1.1072 & -0.7727 & 0.3708 & -0.4272 & 0.7808 & -0.5454 & 0.3625 & -0.2727 \\
Listwise
& \textbf{1.4335} & \textbf{-0.8909} 
& \textbf{1.1978} & \textbf{-0.6909} & \textbf{1.1486} & \textbf{-0.9636} & \textbf{0.9072} & \textbf{-0.6090} \\
\bottomrule
\end{tabular}
}
\caption{
Performance of \method across datasets under Pointwise, Pairwise, and Listwise settings. Columns report Cohen’s $d$ and Spearman’s $\rho$. \textbf{Bold} values indicate the best performance based on the metrics. Full results appear in Appendix~\ref{apx:detailed_results}, Tables~\ref{tbl:estimation_methods_results_d}--\ref{tbl:estimation_methods_results_p}.
}
\label{tbl:estimation_methods_results}
\end{table*}

\subsection{Models}\label{ss:models}
To evaluate the effectiveness of the \method method, we conduct experiments using ten different LLMs\footnote{We utilize the Together AI platform to access these LLMs through an API.}, grouped into five categories based on their parameter sizes. This grouping ensures that our results are not biased toward any particular LLM family or model scale. For a detailed description of the models and their categories, we refer the reader to Appendix~\ref{apx:models_details}.\looseness=-1

\begin{figure}[t]
	\centering
	\includegraphics[width=\columnwidth]{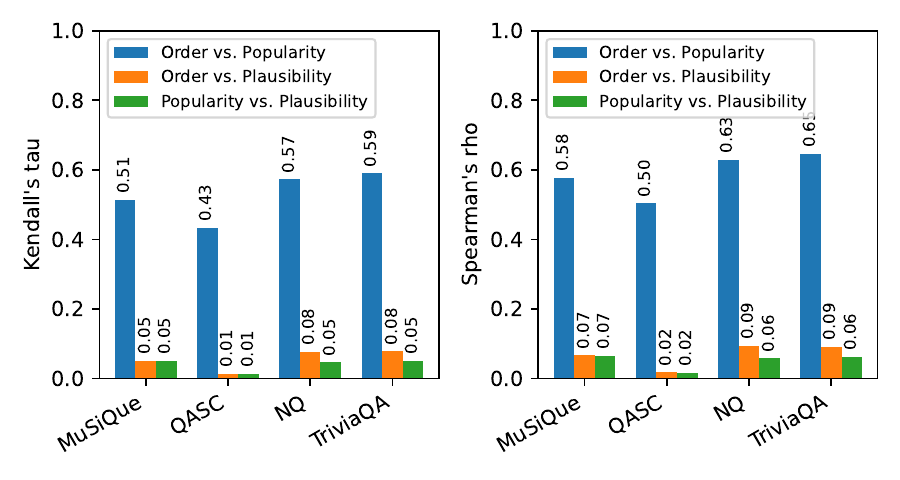}
	\caption{Kendall’s $\tau$ (left chart) and Spearman’s $\rho$ (right chart) correlation coefficients across datasets comparing candidate answer ordering with Popularity, Plausibility scores, and their interplay. 
    % The left chart shows Kendall’s $\tau$ values, and the right chart shows Spearman’s $\rho$ values.
    }
	\label{fig:correlations}
\end{figure}
\subsection{Metrics}\label{ss:metrics}

\subsubsection{Spearman's $\rho$ Correlation}\label{sss:spearman}
We categorize questions by the number of LLMs that successfully answer them. For each category, we compute the average difficulty score of its questions and then calculate Spearman's $\rho$ correlation coefficient between the number of successful LLMs and the corresponding average difficulty. A more negative coefficient indicates stronger alignment between the estimated difficulty scores and actual model performance: as difficulty increases, the number of models able to answer the questions should decrease.\looseness=-1

\subsubsection{Cohen's $d$}\label{sss:cohen_d}
We first define the median of the computed difficulty scores as a threshold \(\tau\), partitioning the set of questions \(Q\) into two equal-sized groups:

{
\footnotesize
\begin{align}
    Q_\text{Easy} &= \{ q \in Q \mid \text{Diff}(q) \le \tau \}, \\
    Q_\text{Hard} &= \{ q \in Q \mid \text{Diff}(q) > \tau \} \notag
\end{align}
}

Here, \(\text{Diff}(q)\) denotes the difficulty score assigned to a question \(q\). The group \(Q_\text{Easy}\) consists of lower-entropy (easier) questions, while \(Q_\text{Hard}\) consists of higher-entropy (harder) questions.

Next, for each question in both groups, we prompt an LLM (selected from the models described in Section~\ref{ss:models}), denoted as $\text{LLM}_{m}$, using the prompt shown in Figure~\ref{fig:qa_prompt} in Appendix~\ref{apx:metrics_details}. The answer generated by $\text{LLM}_{m}$ for a given question \(q\) is denoted as:

{
\footnotesize
\begin{align}
a_{m}(q) = \text{LLM}_m(q)
\end{align}
}

These generated answers are subsequently evaluated for correctness using the \texttt{GPT-Eval} method~\cite{kamalloo-etal-2023-evaluating}, which leverages strong LLMs as the verification system. To mitigate potential evaluator bias arising from reliance on a single model, we employ three distinct LLMs as judges to improve reliability. Specifically, we use GPT-4~\cite{openai2024gpt4technicalreport}, Gemini~2.5 Flash~\cite{2025arXiv250706261C}, and Claude Sonnet~4.5~\cite{anthropic2024sonnet45}. The final correctness label is determined via majority voting across the three judgments. Appendix~\ref{apx:gpt-eval} details our motivation for adopting \texttt{GPT-Eval}, the evaluation procedure, and the exact prompt used.
Formally, for a question \( q \), an LLM-generated answer \( a_m(q) \), and the gold answer \( gt_q \), we define the semantic correctness indicator as:\looseness=-1

{\small
\begin{align}
    \text{Eval}(q, a_m(q), gt_q) = 
    \begin{cases}
    1 & \text{if GPT-Eval is ``Yes''} \\
    0 & \text{if GPT-Eval is ``No''}
    \end{cases}
\end{align}
}

For each group \(G \in \{ Q_\text{Easy}, Q_\text{Hard} \}\), we then define the accuracy of $\text{LLM}_{m}$ as the average \texttt{Eval} correctness score across its answers:

{
\footnotesize
\begin{align}
    \label{eq:accuracy}
    \text{Acc}_{m}(G) = \frac{1}{|G|} \sum_{q \in G} \text{Eval}\left(q, a_m(q)\right)
\end{align}
}

Finally, we use Cohen’s \(d\)~\cite{Cohen2013-zu} to measure the standardized difference between the two groups, providing an interpretable estimate of how clearly the method separates questions based on their difficulties:

{
\footnotesize
\begin{align}
    \label{eq:cohens_d}
    d = \frac{\mu_\text{Easy} - \mu_\text{Hard}}{ \sqrt{ \frac{ \sigma_\text{Easy}^2 + \sigma_\text{Hard}^2 }{2} } }
\end{align}
}

where 

{
\footnotesize
\begin{align}
    \mu_\text{G} = \frac{1}{M} \sum_{m}^{M} \text{Acc}_{m}(Q_\text{G}) \\
    \sigma_\text{G} = \sqrt{ \frac{1}{M} \sum_{m}^{M} \left( \text{Acc}_{m}(Q_\text{G}) - \mu_\text{G} \right)^2 } \notag
\end{align}
}

Where $M$ refers to the set of LLMs described in Section~\ref{ss:models}. By definition, if the mean accuracy for the easy group exceeds that of the hard group, then \(d > 0\); otherwise, \(d < 0\). In other words, the larger Cohen’s $d$ is, the better \method method separates easy and hard questions and vice versa. For additional clarity, Appendix~\ref{apx:cohens_d} includes a worked example, and Table~\ref{tbl:cohens_d_interpretation} summarizes the standard interpretation of Cohen’s $d$ ranges.

%%% EXPERIMENTS AND RESULTS%%%

\begin{table*}
\centering
\resizebox{\textwidth}{!}{%
\begin{tabular}{l|l
  |cc
  |cc
  |cc
  |cc}
\toprule
\textbf{Category} & \textbf{Method} &
\multicolumn{2}{c}{\textbf{MuSiQue}} &
\multicolumn{2}{c}{\textbf{QASC}} &
\multicolumn{2}{c}{\textbf{NQ}} &
\multicolumn{2}{c}{\textbf{TriviaQA}} \\
 &  &
 $d$ & $\rho$ &
 $d$ & $\rho$ &
 $d$ & $\rho$ &
 $d$ & $\rho$ \\
\midrule
\multirow{2}{*}{Readability} 
& Flesch-Kincaid~\cite{rudolf_franz_flesch_1948}
& -0.543  & 0.5545
& 0.1496  & 0.1909
& -0.424  & 0.6363
& -0.2689 & 0.5181 \\
& Gunning-Fog~\cite{Gunning1952}
& -0.3947 & 0.7181
& -0.0944 & -0.0636
& -0.5775 & 0.6272
& -0.0963 & 0.2090 \\
\midrule
\multirow{2}{*}{Prompt-based}
& LLaMA 3.1 8b~\cite{grattafiori2024llama3herdmodels}
& -0.535  & 0.2636
& 0.1065  & -0.1272
& 0.0762  & 0.0818
& 0.361   & -0.2545 \\
& LLaMA 3.3 70b~\cite{grattafiori2024llama3herdmodels}
& 0.2453  & -0.109
& 0.2032  & -0.2909
& 0.0307  & -0.3363
& 0.4566  & -0.4272 \\
\midrule
Popularity
& PopQA~\cite{mallen-etal-2023-trust}
& -0.0275 & 0.2818
& -0.3206 & 0.2727
& 0.1535  & 0.2636
& -0.2702 & 0.1727 \\
\midrule
Retriever-based
& Retrieval Complexity~\cite{gabburo-etal-2024-measuring}
& 0.1284  & -0.3451
& 0.2225  & -0.3126
& 0.2781  & -0.4518
& 0.4394  & -0.5129 \\
\midrule
\multirow{2}{*}{Uncertainty-based}
& LLaMA 3.1 8b~\cite{2024.EDM-posters.90}
& 0.1365  & -0.3815
& 0.1543  & -0.3926
& 0.1556  & -0.5025
& 0.2211  & -0.3121 \\
& LLaMA 3.3 70b~\cite{2024.EDM-posters.90}
& 0.4219  & -0.5518
& 0.2119  & -0.5621
& 0.3265  & -0.5071
& 0.4823  & -0.452 \\
\midrule
\midrule
\multirow{2}{*}{\method}
& Avg-Plausibility
& -0.2242 & 0.0272
& 0.4784 & -0.3
& 0.1869 & -0.2545
& 0.564 & -0.509 \\
& Entropy-Plausibility
& \textbf{1.0888} & \textbf{-0.9001}
& \textbf{0.803} & \textbf{-0.6181}
& \textbf{0.9448} & \textbf{-0.9636}
& \textbf{0.7498} & \textbf{-0.8818} \\
\bottomrule
\end{tabular}
}%
\caption{Comparison of baselines and \method across datasets. \textbf{Bold} indicates the best scores for each dataset. Full results are provided in Appendix~\ref{apx:detailed_results}, Tables~\ref{tbl:model_performance_d}--\ref{tbl:model_performance_p}.}
\label{tbl:model_performance}
\end{table*}

\section{Experiments and Results}\label{s:experiments}

In this section, we present a series of key experiments to evaluate the performance of \method under various settings and to compare it against baseline methods, demonstrating its overall effectiveness. Additional experiments—including generalization (Appendix~\ref{apx:generalization}), and $\alpha$-robustness (Appendix~\ref{apx:alpha_robustness})—are also conducted and detailed in Appendix~\ref{apx:experiments}. Error analysis is also detailed in Appendix~\ref{apx:error_analysis}.

For experiments, we use the optimal values of $\alpha$ and the number of candidate answers, selected via grid search. Specifically, $\alpha$ is searched over the range $[0, 1]$ with a step size of $0.01$, and the number of candidate answers is searched over the range $[1, 20]$ in increments of $1$.

\subsection{Analyzing Popularity Bias}\label{ss:popularity_correlations}

To understand the influence of popularity on LLM-generated candidate answers, we examine three key quantities: (1) the popularity of each candidate answer, (2) the order in which LLMs produce candidate answers, and (3) their plausibility scores. Prior work~\cite{mallen-etal-2023-trust}, has shown that LLMs exhibit popularity bias in their final answers, but it remains unclear whether such bias also appears during the generation of candidate answers.

We evaluate these relationships using Kendall’s $\tau$~\cite{665905b2-6123-3642-832e-05dbc1f48979} and Spearman’s $\rho$~\cite{ca468a70-0be4-389a-b0b9-5dd1ff52b33f}. Specifically, we measure the pairwise correlations between (1) popularity and plausibility scores, (2) the order of candidate answers and plausibility scores, and (3) the order of candidate answers and popularity.

The results, shown in Figure~\ref{fig:correlations}, reveal no significant correlation between plausibility and either popularity or generation order. This confirms that popularity and plausibility reflect fundamentally different properties and should not be used interchangeably. However, we observe a consistent correlation between popularity and the order of generated candidate answers, indicating that LLMs tend to generate more popular answers earlier. This provides clear evidence of a popularity bias already present at the candidate generation stage, which motivates our approach to correcting this effect when computing plausibility scores.

\subsection{Answer Plausibility Estimation Methods}\label{ss:answer_plausibility_estimation_methods}

We investigate three approaches for estimating plausibility scores of candidate answers: \textit{Pointwise}, \textit{Pairwise}, and \textit{Listwise}. We consider these approaches because they are three principled ways to elicit plausibility scores for a pool of candidate answers, each with a different characteristic. \textit{Pointwise} prompts the model separately for each candidate answer, \textit{Pairwise} compares candidates in pairs and aggregates preferences with the Bradley–Terry model~\cite{19ff28b9-64f9-3656-ba40-08326a05748e}, and \textit{Listwise} generates candidates and their plausibility scores jointly. A detailed description of these methods, along with prompts and formal definitions, is provided in Appendices~\ref{apx:answer_plausibility_estimation_methods}, \ref{apx:pointwise_estimation_prompt}, and \ref{apx:pairwise_estimation_prompt}.

For this experiment \textit{only}, we sample 250 questions from each dataset using stratified sampling~\cite{ARNAB2017213} based on question types. Running this experiment on the full datasets would be computationally prohibitive; for instance, the \textit{Pairwise} approach alone would require approximately 3.2 million prompts. Sampling 250 questions per dataset therefore enables a fair and tractable comparison of the different approaches under a fixed computational budget.

As shown in Table~\ref{tbl:estimation_methods_results},
among the three estimation strategies, the \textit{Listwise} approach achieves the strongest overall performance. Beyond accuracy, it is also the most efficient, both in terms of the number of prompts and the average output length, requiring only a \textit{single} prompt with complexity $\mathcal{O}(1)$, compared to $\mathcal{O}(n)$ for \textit{Pointwise} and $\mathcal{O}(n^2)$ for \textit{Pairwise}, as reported in Table~\ref{tbl:scenarios_comparison} in Appendix~\ref{apx:answer_plausibility_estimation_methods}. Taken together, these findings establish \textit{Listwise} as the most effective and practical method for plausibility-based difficulty estimation.

\begin{table}[t]
\centering
\resizebox{\columnwidth}{!}{%
\begin{tabular}{c@{\hspace{45pt}} c c @{\hspace{45pt}} c c}
\toprule
\textbf{Evaluator} & \textbf{Age} & \textbf{Education Level} & \textbf{Easy} & \textbf{Hard}\\
\midrule
Evaluator 1 & 18 & High School & 0.65 & 0.68 \\
Evaluator 2 & 36 & High School & 0.72 & 0.70  \\
Evaluator 3 & 30 & Bachelor's Degree & 0.68 & 0.82 \\
Evaluator 4 & 33 & PhD & 0.72 & 0.70 \\
Evaluator 5 & 40 & MSc & 0.68 & 0.75 \\
Evaluator 6 & 35 & PhD & 0.70 & 0.80 \\
\midrule
\textbf{Mean Accuracy} & \textbf{} & \textbf{} & \textbf{0.68} & \textbf{0.74} \\
\bottomrule
\end{tabular}%
}
\caption{Per-evaluator accuracy in detecting the correct difficulty.}
\label{tbl:evaluator_accuracy_by_category}
\end{table}

\begin{figure}[t]
	\centering
	\includegraphics[width=0.8\columnwidth]{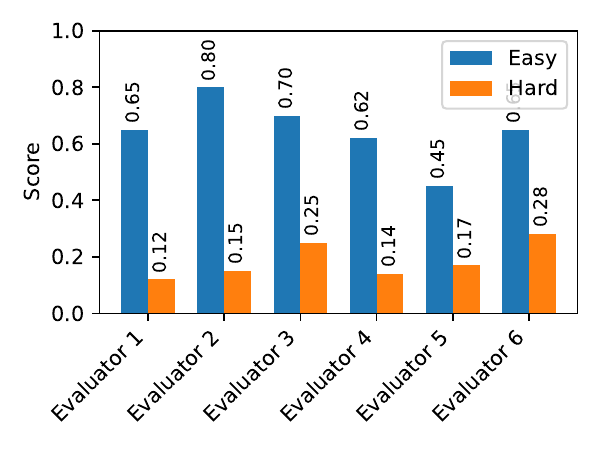}
\caption{Results of human evaluation. The \textcolor{blue}{blue} columns indicate each evaluator’s accuracy on easy questions, while the \textcolor{orange}{orange} columns represent their accuracy on hard questions.}
	\label{fig:human_evaluation}
\end{figure}

\subsection{\method Performance}\label{ss:model_performance}
Based on the closed relevant work discussed in Appendix~\ref{s:related_works}, we apply the following baselines:

\paragraph{Readability}
We include two traditional readability measures: Flesch-Kincaid~\cite{rudolf_franz_flesch_1948}, based on average sentence length and syllable count, and Gunning-Fog~\cite{Gunning1952}, which emphasizes complex word frequency and sentence length.

\paragraph{Prompt-based}

We prompt LLaMA 3.3 70B and LLaMA 3.1 8B with: \textit{Please rate the difficulty of the question from 0 to 100, and respond with a number only}, to directly estimate difficulty.

\paragraph{Popularity}
We adopt PopQA~\cite{mallen-etal-2023-trust}, which uses normalized Wikipedia page views as a proxy for question difficulty.

\paragraph{Retriever-based}
We include the method from~\citet{gabburo-etal-2024-measuring}, which estimates difficulty based on answerability and completeness of passages retrieved by a retriever.

\paragraph{Uncertainty-based}
We include the method from~\citet{2024.EDM-posters.90}, which estimates difficulty from the QA loss of LLMs, capturing model uncertainty when generating the correct answer.

\paragraph{\method (ours)}
We consider two variants of our approach: \textit{Avg-Plausibility}, which computes the difficulty score as the mean of the plausibility scores across candidate answers, and \textit{Entropy-Plausibility}, which defines the difficulty score as the entropy of the plausibility distribution. We adopt the \textit{Listwise} scenario as the primary configuration for \method, as Section~\ref{ss:answer_plausibility_estimation_methods} shows that this scenario achieves the best performance.

Table~\ref{tbl:model_performance} shows that estimating difficulty using the entropy of plausibility outperforms using the average plausibility. This indicates that entropy better captures the notion of difficulty by reflecting the uncertainty in the plausibility distribution. Furthermore, compared with all baselines, \method achieves the highest performance, demonstrating its effectiveness in accurately modeling question difficulty.
The second-best performance is obtained by the uncertainty-based method, followed by the retriever-based, which outperforms the prompt-based approach. These results indicate that estimating question difficulty is a non-trivial problem: directly querying an LLM or relying on simple heuristic computations is insufficient. Instead, more principled and novel modeling approaches are required to achieve strong performance.

\subsection{Human Evaluation}\label{apx:human_evaluation}
\begin{figure}
	\centering
	\includegraphics[width=\columnwidth]{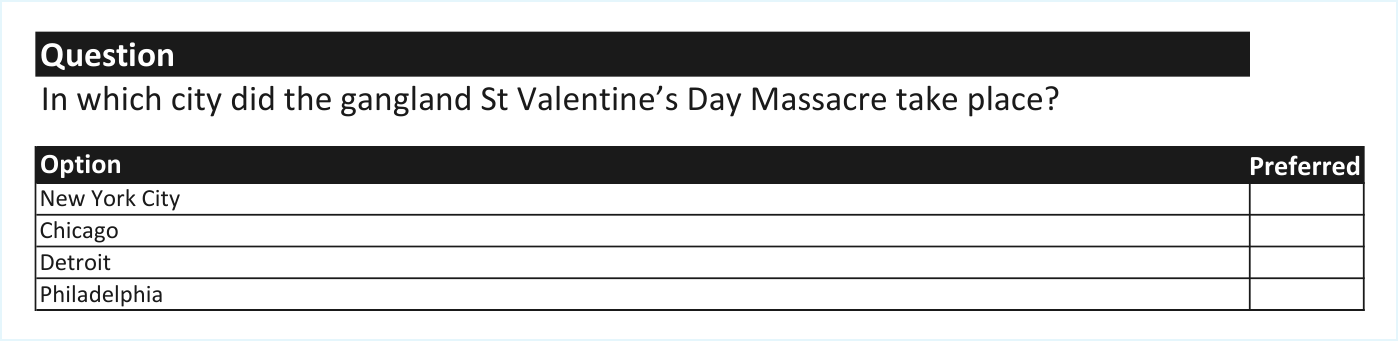}
	\caption{The template of the Excel sheet of questions for MCQA evaluation}
	\label{fig:excel_template_mcqa}
\end{figure}
\begin{figure}[t!]
	\centering
	\includegraphics[width=\columnwidth]{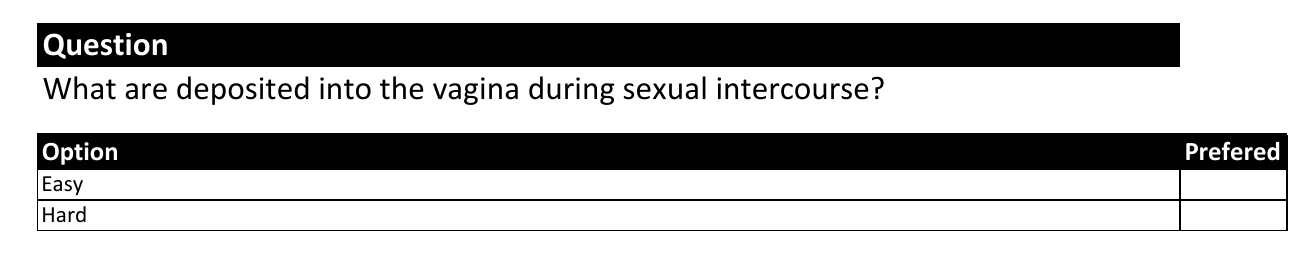}
	\caption{The template of the Excel sheet of questions for the Difficulty Ratings evaluation}
	\label{fig:excel_template_score}
\end{figure}
\begin{table}[]
\resizebox{\columnwidth}{!}{%
\begin{tabular}{@{}llcccc@{}}
\toprule
\textbf{Category} &
  \textbf{Method} &
  \textbf{MuSiQue} &
  \textbf{QASC} &
  \textbf{NQ} &
  \textbf{TriviaQA} \\ \midrule
\multicolumn{1}{l|}{\multirow{2}{*}{Readability}} &
  \multicolumn{1}{l|}{Flesch-Kincaid} &
  -0.543 &
  0.1496 &
  -0.424 &
  -0.2689 \\
\multicolumn{1}{l|}{} &
  \multicolumn{1}{l|}{Gunning-Fog} &
  -0.3947 &
  -0.0944 &
  -0.5775 &
  -0.0963 \\ \midrule
\multicolumn{1}{l|}{\multirow{2}{*}{Prompt-based}} &
  \multicolumn{1}{l|}{LLaMA 3.1 8b} &
  -0.535 &
  0.1065 &
  0.0762 &
  0.361 \\
\multicolumn{1}{l|}{} &
  \multicolumn{1}{l|}{LLaMA 3.3 70b} &
  0.2453 &
  0.2032 &
  0.0307 &
  0.4566 \\ \midrule
\multicolumn{1}{l|}{Popularity} &
  \multicolumn{1}{l|}{PopQA} &
  -0.0275 &
  -0.3206 &
  0.1535 &
  -0.2702 \\ \midrule
\multicolumn{1}{l|}{Retriever-based} &
  \multicolumn{1}{l|}{Retrieval Complexity} &
  0.1284 &
  0.2225 &
  0.2781 &
  0.4394 \\ \midrule
\multicolumn{1}{l|}{\multirow{2}{*}{Uncertainty-based}} &
  \multicolumn{1}{l|}{LLaMA 3.1 8b} &
  0.1365 &
  0.1543 &
  0.1556 &
  0.2211 \\
\multicolumn{1}{l|}{} &
  \multicolumn{1}{l|}{LLaMA 3.3 70b} &
  0.4219 &
  0.2119 &
  0.3265 &
  0.4823 \\ \midrule \midrule
\multicolumn{1}{l|}{\multirow{2}{*}{\method}} &
\multicolumn{1}{l|}{Without gold answer} &
   {\ul 0.8325} &
   {\ul 0.5144}&
   {\ul 0.6319} &
   {\ul 0.6647}\\
\multicolumn{1}{l|}{} &
  \multicolumn{1}{l|}{With gold answer} &
  \textbf{1.0888} &
  \textbf{0.803} &
  \textbf{0.9448} &
  \textbf{0.7498} \\ \bottomrule
\end{tabular}%
}
\caption{Performance of \method with and without providing gold answer as input, evaluated using Cohen’s $d$ and compared with baseline models. \textbf{Bold} values indicate the best performance for each dataset, and {\ul underlined} values indicate the second-best.}

\label{tbl:ablation_without_gold_answers}
\end{table}

To assess the alignment of \method with human judgments, we conduct two complementary experiments. First, we convert a sampled set of questions into a multiple-choice QA format and measure human accuracy to examine whether questions labeled as difficult by \method are also harder for humans. Second, we ask human annotators to directly judge question difficulty by choosing whether each question is easy or hard, allowing us to compare \method’s difficulty labels with human binary difficulty judgments.

To prepare the evaluation set, we label all questions in each dataset using \method and then sample 30 questions from each difficulty group (Easy and Hard) per dataset, resulting in 60 questions per dataset and 240 questions in total. Six human participants with diverse backgrounds took part in the evaluation. 

\subsubsection{Multiple-Choice QA}\label{ss:multichoice}

In this experiment, each sampled question is converted into a multiple-choice format with four answer options presented in random order. One option is the ground-truth answer, and the remaining three are selected from the most plausible candidate answers generated for that question. Figure~\ref{fig:excel_template_mcqa} illustrates the Excel template used for collecting annotators’ responses.

Figure~\ref{fig:human_evaluation} shows the results of the human MCQA evaluation. As expected, all evaluators achieved higher accuracy on questions labeled as Easy than on those labeled as Hard. This demonstrates that \method effectively captures question difficulty in a manner consistent with human performance.

\subsubsection{Question Difficulty Ratings}\label{ss:question_difficulty}

In the second experiment, we evaluate the extent to which human annotators agree with the difficulty labels assigned by \method. Human evaluators are then asked to judge the difficulty of each sampled question by choosing whether it is easy or hard. Figure~\ref{fig:excel_template_score} illustrates the annotation template used to collect these judgments.

To quantify how well evaluators identify the intended difficulty level, we compute accuracy by comparing human judgments against the \method-assigned labels for easy and hard questions. The results are summarized in Table~\ref{tbl:evaluator_accuracy_by_category}. As shown, evaluators achieve higher accuracy on hard questions, indicating that harder questions are more consistently recognized by humans, whereas easy questions exhibit greater subjectivity in perceived difficulty.

Overall, the per-difficulty accuracy analysis results confirm that \method produces difficulty estimates that align well with human intuition and are robust across different types of questions.

\subsection{Ablation Study}\label{ss:ablation_study}
In this experiment, we evaluate the performance of \method by examining three factors: (1) the effect of removing the gold answer, (2) the impact of excluding the debiasing component, and (3) the influence of using different LLMs.

\begin{table}[]
\resizebox{\columnwidth}{!}{%
\begin{tabular}{@{}llcccc@{}}
\toprule
\textbf{Category} &
  \textbf{Method} &
  \textbf{MuSiQue} &
  \textbf{QASC} &
  \textbf{NQ} &
  \textbf{TriviaQA} \\ \midrule
\multicolumn{1}{l|}{\multirow{2}{*}{Readability}} &
  \multicolumn{1}{l|}{Flesch-Kincaid} &
  -0.543 &
  0.1496 &
  -0.424 &
  -0.2689 \\
\multicolumn{1}{l|}{} &
  \multicolumn{1}{l|}{Gunning-Fog} &
  -0.3947 &
  -0.0944 &
  -0.5775 &
  -0.0963 \\ \midrule
\multicolumn{1}{l|}{\multirow{2}{*}{Prompt-based}} &
  \multicolumn{1}{l|}{LLaMA 3.1 8b} &
  -0.535 &
  0.1065 &
  0.0762 &
  0.361 \\
\multicolumn{1}{l|}{} &
  \multicolumn{1}{l|}{LLaMA 3.3 70b} &
  0.2453 &
  0.2032 &
  0.0307 &
  0.4566 \\ \midrule
\multicolumn{1}{l|}{Popularity} &
  \multicolumn{1}{l|}{PopQA} &
  -0.0275 &
  -0.3206 &
  0.1535 &
  -0.2702 \\ \midrule
\multicolumn{1}{l|}{Retriever-based} &
  \multicolumn{1}{l|}{Retrieval Complexity} &
  0.1284 &
  0.2225 &
  0.2781 &
  0.4394 \\ \midrule
\multicolumn{1}{l|}{\multirow{2}{*}{Uncertainty-based}} &
  \multicolumn{1}{l|}{LLaMA 3.1 8b} &
  0.1365 &
  0.1543 &
  0.1556 &
  0.2211 \\
\multicolumn{1}{l|}{} &
  \multicolumn{1}{l|}{LLaMA 3.3 70b} &
  0.4219 &
  0.2119 &
  0.3265 &
  0.4823 \\ \midrule \midrule
\multicolumn{1}{l|}{\multirow{2}{*}{\method}} &
  \multicolumn{1}{l|}{Without debiasing} &
  {\ul 0.894} &
  {\ul 0.5614} &
  {\ul 0.88} &
  {\ul 0.6511} \\
\multicolumn{1}{l|}{} &
  \multicolumn{1}{l|}{With debiasing} &
  \textbf{1.0888} &
  \textbf{0.803} &
  \textbf{0.9448} &
  \textbf{0.7498} \\ \bottomrule
\end{tabular}%
}
\caption{Performance of \method with and without popularity debiasing, evaluated using Cohen’s $d$ and compared with baseline models. \textbf{Bold} values indicate the best performance for each dataset, and {\ul underlined} values indicate the second-best.}

\label{tbl:ablation_without_debiasing}
\end{table}

\paragraph{Without Gold Answers}
We evaluate \method without providing the gold answer as input. Figure~\ref{fig:prompt_golden_answer_inclusion} in Appendix~\ref{apx:gold_answer_inclusion} shows the prompt used in this setting. Here, the LLM generates candidate answers without knowing the gold answer, reflecting realistic scenarios where gold answers are unavailable. As shown in Table~\ref{tbl:ablation_without_gold_answers}, \method still outperforms all baselines, although performance decreases compared to the setting with a gold answer. This is expected, as access to the gold answer allows the LLM to generate higher-quality candidate answers. Appendix~\ref{apx:gold_answer_inclusion} provides a detailed explanation of why including the gold answer leads to better performance. Overall, these results demonstrate that \method remains effective even when gold answers are missing and does not overly rely on them.

\paragraph{Without Debiasing}
We evaluate \method without the popularity debiasing component. In this variant, we compute the difficulty score directly from the entropy of the raw plausibility distribution. Table~\ref{tbl:ablation_without_debiasing} shows that \method continues to outperform all baselines even without debiasing, though performance is lower than the full method. This aligns with our earlier findings in Section~\ref{ss:popularity_correlations}, where we observed a correlation between candidate answer order and popularity; removing the debiasing step can therefore lead to less reliable plausibility estimates. Nonetheless, the strong performance of this variant indicates that \method remains robust in domains where Wikipedia page views may not accurately capture popularity (e.g., medicine, finance, or specialized technical fields).

\begin{table}[]
\resizebox{\columnwidth}{!}{%
\begin{tabular}{@{}llcccc@{}}
\toprule
\textbf{Category} &
  \textbf{Method} &
  \textbf{MuSiQue} &
  \textbf{QASC} &
  \textbf{NQ} &
  \textbf{TriviaQA} \\ \midrule
\multicolumn{1}{l|}{\multirow{2}{*}{Readability}} &
  \multicolumn{1}{l|}{Flesch-Kincaid} &
  -0.543 &
  0.1496 &
  -0.424 &
  -0.2689 \\
\multicolumn{1}{l|}{} &
  \multicolumn{1}{l|}{Gunning-Fog} &
  -0.3947 &
  -0.0944 &
  -0.5775 &
  -0.0963 \\ \midrule
\multicolumn{1}{l|}{\multirow{2}{*}{Prompt-based}} &
  \multicolumn{1}{l|}{LLaMA 3.1 8b} &
  -0.535 &
  0.1065 &
  0.0762 &
  0.361 \\
\multicolumn{1}{l|}{} &
  \multicolumn{1}{l|}{LLaMA 3.3 70b} &
  0.2453 &
  0.2032 &
  0.0307 &
  0.4566 \\ \midrule
\multicolumn{1}{l|}{Popularity} &
  \multicolumn{1}{l|}{PopQA} &
  -0.0275 &
  -0.3206 &
  0.1535 &
  -0.2702 \\ \midrule
\multicolumn{1}{l|}{Retriever-based} &
  \multicolumn{1}{l|}{Retrieval Complexity} &
  0.1284 &
  0.2225 &
  0.2781 &
  0.4394 \\ \midrule
\multicolumn{1}{l|}{\multirow{2}{*}{Uncertainty-based}} &
  \multicolumn{1}{l|}{LLaMA 3.1 8b} &
  0.1365 &
  0.1543 &
  0.1556 &
  0.2211 \\
\multicolumn{1}{l|}{} &
  \multicolumn{1}{l|}{LLaMA 3.3 70b} &
  0.4219 &
  0.2119 &
  0.3265 &
  {\ul 0.4823} \\ \midrule \midrule
\multicolumn{1}{l|}{\multirow{3}{*}{\method}} &
  \multicolumn{1}{l|}{Qwen 2.5 7b} &
  {\ul 0.8434} &
  0.1465 &
  0.2465 &
  0.3162 \\
\multicolumn{1}{l|}{} &
  \multicolumn{1}{l|}{LLaMA 3.1 8b} &
  0.5467 &
  {\ul 0.2484} &
  {\ul 0.3886} &
  0.3481 \\
\multicolumn{1}{l|}{} &
  \multicolumn{1}{l|}{LLaMA 3.3 70b} &
  \textbf{1.0888} &
  \textbf{0.803} &
  \textbf{0.9448} &
  \textbf{0.7498} \\ \bottomrule
\end{tabular}%
}
\caption{Performance of different LLMs as the core of \method, evaluated using Cohen’s $d$ and compared with baseline models. \textbf{Bold} values indicate the best performance for each dataset, and {\ul underlined} values indicate the second-best. Full results in Appendix~\ref{apx:detailed_results}, Tables~\ref{tbl:core_results_d}--\ref{tbl:core_results_p}.}

\label{tbl:ablation_influence_of_llm}
\end{table}

\paragraph{Influence of LLM Choice}
We evaluate how different LLMs influence the performance of \method by using Qwen 2.5 7B~\cite{qwen2025qwen25technicalreport}, LLaMA 3.1 8B~\cite{grattafiori2024llama3herdmodels}, and LLaMA 3.3 70B~\cite{grattafiori2024llama3herdmodels} as the generation core. Table~\ref{tbl:ablation_influence_of_llm} shows that \method performs strongly even with smaller models, consistently surpassing baseline methods. The only exception is the TriviaQA dataset, where Uncertainty-based scoring with LLaMA 3.3 70B slightly outperforms \method—an expected outcome given its use of a significantly larger and more capable model. For the other datasets, LLaMA 3.1 8B and Qwen 2.5 7b achieve the second-best results across baselines, demonstrating that \method remains effective even with relatively small LLMs. As anticipated, larger and more capable models yield stronger results overall, reflecting their superior generation quality—a trend well documented across NLP tasks. These findings indicate that \method is not only effective with powerful LLMs but also practical in resource-constrained settings, supporting its usability and reproducibility in real-world applications.

\paragraph{Summary of Ablation Findings.}
Overall, \method remains robust across all ablation settings. While larger LLMs and access to gold answers improve performance, the method continues to outperform all baselines even with smaller models and without gold answers, confirming its practicality in realistic scenarios. Popularity debiasing further enhances performance, but its removal does not break the method, indicating that debiasing is a beneficial refinement rather than a strict requirement. A paired t-test comparing the \textit{With-Debiasing} and \textit{Without-Debiasing} variants shows statistically significant improvements in difficulty separation (p = 0.0356, t = 3.6474, Cohen’s $d$), reinforcing the effectiveness of the debiasing component.

\section{Conclusion}\label{s:conclusion}
We introduced Q-DAPS, an LLM-oriented method for estimating question difficulty based on the entropy of answer plausibility scores. By modeling how plausible competing candidate answers appear to an LLM and mitigating popularity bias, Q-DAPS captures difficulty in a way that reflects underlying reasoning uncertainty rather than surface-level textual features.
Experiments across four QA benchmarks show that Q-DAPS consistently outperforms readability-based, popularity-based, retriever-based, prompt-based, and uncertainty-based baselines in separating easy and hard questions. Ablation studies indicate that the method remains robust across plausibility estimation paradigms, model sizes, and realistic settings without gold answers or popularity debiasing. Human evaluations further confirm strong alignment with human judgments of difficulty.
Overall, Q-DAPS provides an interpretable and scalable method for question difficulty estimation in LLM-based QA systems, supporting applications such as hallucination risk estimation, adaptive model selection, question routing, and safeguard triggering. Future work will explore extensions to multilingual and multimodal settings, as well as tighter integration of difficulty estimates into adaptive and safety-aware QA pipelines.

\section*{Limitations} \label{s:limitations}  
The \method approach and its associated experiments have some limitations:

\begin{itemize}
    \item \textbf{Question Type:} \method is especially suited to question types for which a compact set of plausible candidates can be constructed, including entity, temporal, numeric, boolean, categorical/label selection, and multiple-choice (native or synthesized).
    These formats are common in enterprise and education contexts, making \method broadly applicable without requiring extensive labeled data.
    
    \item \textbf{Language Scope:} Our study focuses solely on English-language questions. As a result, the methods and findings may not directly transfer to other languages. Further work is needed to assess the approach's effectiveness in multilingual or low-resource language settings.
    
    \item \textbf{Model Dependency:} The pipeline depends on LLMs, whose behavior, performance, and biases influence the outcomes. These models may encode societal or cultural biases and may behave inconsistently across different contexts.
\end{itemize}

\section*{Ethical Considerations} \label{s:ethics}  
This study employs GPT models licensed under OpenAI and Apache 2.0, as well as LLaMA models governed by Meta’s LLAMA Community License Agreement. We fully comply with all licensing terms. The datasets used are sourced from platforms permitting academic use. To support reproducibility and further research, we release all study materials under the MIT license. Throughout the project, we have ensured that data handling, model usage, and result dissemination conform to relevant ethical standards and legal requirements.

\section*{Acknowledgments} \label{s:acknowledgments}  
The computational results presented here have been achieved (in part) using the LEO HPC infrastructure of the University of Innsbruck and the Austrian Scientific Computing (ASC) infrastructure.

\bibliography{custom}
\bibliographystyle{acl_natbib}

\clearpage

\appendix

\section{Related Works}\label{s:related_works}

\begin{table*}[t]
\centering
\resizebox{\linewidth}{!}{%
\begin{tabular}{l|l|cccccc}
\toprule
\textbf{Perspective} & \textbf{Representative Approaches} & \textbf{Surface-level} & \textbf{Retriever-based} & \textbf{Uncertainty-based} & \textbf{LLM-oriented} & \textbf{Popularity Debiasing} & \textbf{Interpretable} \\
\midrule
Readability &
\makecell[l]{Traditional readability formulas~\cite{Gunning1952,rudolf_franz_flesch_1948} \\ 
Hybrid ML-based methods~\cite{liu-lee-2023-hybrid} \\ 
Prompt-based estimation~\cite{naous-etal-2024-readme}} &
\checkmark & $\times$ & $\times$ & $\times$ & $\times$ & $\times$ \\
\midrule
Retrieval-based &
\makecell[l]{BM25 retrieval with top-10 documents~\cite{jamshid_wikihint} \\ 
Answerability and completeness metrics~\cite{gabburo-etal-2024-measuring}} &
$\times$ & \checkmark & $\times$ & $\times$ & $\times$ & $\times$ \\
\midrule
LLM-based &
\makecell[l]{Model confidence and uncertainty~\cite{2024.EDM-posters.90} \\ 
Uncertainty for MCQ difficulty~\cite{zotos2025doubtfulohdifficultthen}} &
$\times$ & $\times$ & \checkmark & $\times$ & $\times$ & $\times$ \\
\midrule
\textbf{\method (ours)} &
\makecell[l]{Comparative plausibility over candidate answers \\ with popularity debiasing} &
$\times$ & $\times$ & \checkmark & \checkmark & \checkmark & \checkmark \\
\bottomrule
\end{tabular}
}
\caption{Different perspectives on question difficulty estimation, with representative approaches, feature coverage, and our proposed \method, which is the most complete.}
\label{tbl:perspectives_question_difficulty}
\end{table*}

Questions can be typically classified according to various criteria, including \textit{Answer Type}, \textit{Semantic Type}, \textit{Question Structure}, \textit{Intent}, and \textit{Difficulty}.

\textbf{Answer Type} refers to the nature of the expected response, such as factoid~\cite{kalouli-etal-2021-really} or non-factoid~\cite{Breja02092022}.
\textbf{Semantic Type} considers the kind of information sought, for example, entities~\cite{sciavolino-etal-2021-simple}, definitions~\cite{hildebrandt-etal-2004-answering}, or causal explanations~\cite{bondarenko-etal-2022-causalqa}.
\textbf{Question Structure} describes the grammatical form of the question, including yes/no~\cite{clark-etal-2019-boolq}, wh-questions~\cite{kalbaliyev-sirts-2022-narrative}, and multiple-choice questions~\cite{manakul-etal-2023-mqag}.
\textbf{Intent} captures the purpose of the question, for example to seek factual information~\cite{abujabal-etal-2019-comqa} or to request clarification~\cite{chi2024clarinetaugmentinglanguagemodels}.
Finally, \textbf{Difficulty} measures how challenging a question is to answer, ranging from straightforward fact-based queries~\cite{pado-2017-question} to complex reasoning problems~\cite{xu-etal-2024-adaption}.

\subsection{Difficulty}\label{ss:difficulty}

Question difficulty can be assessed from different perspectives. One common perspective focuses on \textit{readability}. Traditional readability-based approaches rely on lexical and syntactic features~\cite{Gunning1952,rudolf_franz_flesch_1948} but often fail to capture deeper semantic aspects. To address this, some studies apply machine learning techniques~\cite{liu-lee-2023-hybrid}, while others leverage LLMs with prompt-based methods to estimate difficulty~\cite{naous-etal-2024-readme}.

Another perspective involves \textit{retrieval-based} methods, which evaluate difficulty according to how easily relevant passages can be retrieved. For instance,~\citet{jamshid_wikihint, mozafari-etal-2024-exploring} use BM25~\cite{robertson2009probabilistic} to retrieve the top-10 documents and classify a question as easy if one of these documents contained a correct answer. Similarly,~\citet{gabburo-etal-2024-measuring} expand this idea by incorporating notions of answerability and completeness.

A further line of work focuses on \textit{LLM-based} metrics, such as model confidence and uncertainty~\cite{2024.EDM-posters.90, jain2025exploringpotentiallargelanguage}. For example,~\citet{zotos2025doubtfulohdifficultthen} use model uncertainty to estimate item difficulty for multiple-choice questions. However, these approaches have largely targeted question difficulty from a human perspective, rather than explicitly addressing how difficult a question is for an LLM to answer.

Table~\ref{tbl:perspectives_question_difficulty} summarizes existing perspectives on question difficulty estimation, representative approaches, and their limitations.

\subsection{Entropy in NLP}\label{ss:entropy}
Entropy has been applied in various NLP and Data Analysis tasks~\cite{byteplus_entropy2025, e26121126}. For example,~\citet{KHURANA2022115820} used entropy to measure sentence informativeness in extractive summarization, employing Non-negative Matrix Factorization to derive probability distributions over terms, topics, and sentences. Likewise,~\citet{li2023unlockingcontextconstraintsllms} introduced Selective Context, a technique that filters out less informative content based on entropy-derived self-information, improving the efficiency of LLMs.

Recent research has also examined plausibility-based scoring of LLM responses, such as Verbalized Confidence~\cite{tian-etal-2023-just} and CONQORD~\cite{tao-etal-2024-trust}, but these approaches focus on confidence estimates for correct answers only. None of them examine candidate answers and their plausibility scores.

\subsection{Popularity Bias}\label{ss:popularity_bias}
Popularity bias is a well-documented challenge in recommender systems, where algorithms tend to overrepresent highly popular items while underrepresenting niche or less-known items~\cite{Klimashevskaia2024,YALCIN2022103100}. In recommender systems, this can lead to unfair or skewed exposure of content.

In the context of LLMs,~\citet{mallen-etal-2023-trust} showed that these models often struggle with less popular factual knowledge, indicating that popularity bias can influence their generated outputs. Similarly,~\citet{ni2025knowledgepopularityinfluencesenhances} demonstrated that LLMs tend to perform better, exhibit higher confidence, and more accurately perceive their knowledge boundaries when dealing with more popular knowledge, highlighting a strong correlation between knowledge popularity and QA performance. However, no previous work has investigated how popularity bias impacts question difficulty estimation for LLMs or proposed debiasing methods in this context.

\newpage

\section{Case Study}\label{apx:case_study}
% \begin{table*}[t]
% \centering
% \resizebox{\textwidth}{!}{%
% \begin{tabular}{@{}clcccc@{}}
% \toprule
% \textbf{\#} & \textbf{Candidate Answer} & \textbf{Plausibility Score} & \textbf{Wikipedia Popularity} & \textbf{Debiased Plausibility} & \textbf{Normalized Debiased Plausibility} \\
% \midrule
% 1  & Cologne & 80 & 1.000 & 40.000 & 0.054 \\
% 2  & Dortmund & 70 & 0.545 & 50.925 & 0.069 \\
% 3  & Mülheim an der Ruhr & 58 & 0.064 & 56.144 & 0.076 \\
% 4  & Bottrop & 55 & 0.049 & 53.653 & 0.073 \\
% 5  & Bochum & 48 & 0.198 & 43.248 & 0.059 \\
% 6  & Marl & 15 & 0.152 & 13.860 & 0.019 \\
% 7  & Oberhausen & 50 & 0.086 & 47.850 & 0.065 \\
% 8  & Essen & 65 & 0.383 & 52.553 & 0.071 \\
% 9  & Wesel & 38 & 0.064 & 36.784 & 0.050 \\
% 10 & Voerde & 32 & 0.007 & 31.888 & 0.043 \\
% 11 & Recklinghausen & 40 & 0.042 & 39.160 & 0.053 \\
% 12 & Moers & 28 & 0.038 & 27.468 & 0.037 \\
% 13 & Castrop-Rauxel & 42 & 0.029 & 41.391 & 0.056 \\
% 14 & Rheinberg & 22 & 0.024 & 21.736 & 0.030 \\
% 15 & Dinslaken & 35 & 0.016 & 34.720 & 0.047 \\
% 16 & Waltrop & 20 & 0.010 & 19.900 & 0.027 \\
% 17 & Haltern am See & 18 & 0.015 & 17.865 & 0.024 \\
% 18 & Kamp-Lintfort & 30 & 0.016 & 29.760 & 0.040 \\
% 19 & Neukirchen-Vluyn & 25 & 0.008 & 24.900 & 0.034 \\
% 20 & Gelsenkirchen & 60 & 0.258 & 52.260 & 0.071 \\
% \bottomrule
% \end{tabular}%
% }
% \caption{Candidate answers with their plausibility scores, Wikipedia popularity, debiased plausibility, and normalized debiased plausibility for the question \textit{Industrial city in Germany on the Rhine-Herne Canal?}.}
% \label{tbl:case_study_candidate_answers}
% \end{table*}

\begin{table*}[t]
\centering
\resizebox{\textwidth}{!}{%
\begin{tabular}{@{}clcccc@{}}
\toprule
\textbf{\#} & \textbf{Candidate Answer} & \textbf{Plausibility Score} & \textbf{Wikipedia Popularity} & \textbf{Debiased Plausibility} & \textbf{Normalized Debiased Plausibility} \\
\midrule
1  & Sigmund Freud      & 10 & 1.000 & 5.000  & 0.009 \\
2  & Neal Miller        & 20 & 0.029 & 19.710 & 0.037 \\
3  & Jean Piaget        & 5  & 0.785 & 3.038  & 0.006 \\
4  & \underline{Ernst Hilgard}      & \underline{25} & \underline{0.176} & \underline{22.800} & \underline{0.043} \\
5  & Ivan Pavlov        & 20 & 0.637 & 13.630 & 0.026 \\
6  & Carl Rogers        & 10 & 0.462 & 7.690  & 0.014 \\
7  & Lev Vygotsky       & 5  & 0.444 & 3.890  & 0.007 \\
8  & Albert Bandura     & 70 & 0.373 & 56.945 & 0.107 \\
9  & Harry Harlow       & 25 & 0.275 & 21.563 & 0.040 \\
10 & Edward Thorndike   & 60 & 0.198 & 54.060 & 0.101 \\
11 & Konrad Lorenz      & 30 & 0.180 & 27.300 & 0.051 \\
12 & B.F. Skinner       & 80 & 0.890 & 44.400 & 0.083 \\
13 & Gordon Allport     & 15 & 0.134 & 13.995 & 0.026 \\
14 & Walter Mischel     & 35 & 0.062 & 33.915 & 0.064 \\
15 & Edward Tolman      & 40 & 0.056 & 38.880 & 0.073 \\
16 & Clark Hull         & 50 & 0.050 & 48.750 & 0.092 \\
17 & Julian Rotter      & 45 & 0.030 & 44.325 & 0.083 \\
18 & Abraham Maslow     & 5  & 0.730 & 3.175  & 0.006 \\
19 & O. Hobart Mowrer   & 30 & 0.014 & 29.790 & 0.056 \\
20 & Kenneth Spence     & 40 & 0.007 & 39.860 & 0.075 \\
\bottomrule
\end{tabular}%
}
\caption{Candidate answers with their plausibility scores, Wikipedia popularity, debiased plausibility, and normalized debiased plausibility for the question \textit{Who is regarded as the father of modern behaviorism?}. The \underline{candidate answer} indicates the example being used in the case study.}

\label{tbl:case_study_candidate_answers}
\end{table*}

In this section, we present an example that illustrates how the \method method works by following the stages explained in Section~\ref {s:method}. The question we use is \textit{Who is regarded as the father of modern behaviorism?} from NQ~\cite{kwiatkowski-etal-2019-natural} dataset that its answer is \textit{John B. Watson}. So, in the following, we explain each stage separately.

\subsection{Candidate Generation}

In this stage, we prompt LLaMA 3.3~\cite{grattafiori2024llama3herdmodels} using the template shown in Figure~\ref{fig:listwise_prompt} to generate $20$ candidate answers along with plausibility scores in the \textit{Candidate Generation} component. The resulting outputs are presented in Table~\ref{tbl:case_study_candidate_answers}, showing all $20$ candidate answers and their plausibility scores. After generation, these candidates are passed to the \textit{Validation} component to ensure their quality through the following checks:

\begin{enumerate}
    \item \textbf{No Duplicates:} We compare the generated candidate answers using a semantic equality function. As shown in Table~\ref{tbl:case_study_candidate_answers}, there are no duplicate answers.
    \item \textbf{Range of Plausibility Scores:} We verify that all plausibility scores fall within the valid range of $0$ to $100$. Table~\ref{tbl:case_study_candidate_answers} confirms this condition is satisfied.
    \item \textbf{Number of Candidate Answers:} We check that the LLM produced exactly $20$ candidate answers as requested. Table~\ref{tbl:case_study_candidate_answers} confirms this requirement is met.
\end{enumerate}

\subsection{Popularity Debiasing}
In this stage, we employ the HintEval toolkit~\cite{mozafari2025hinteval} to compute the popularity\footnote{HintEval refers to this as \textit{Familiarity}.} of each candidate answer. The \textit{Popularity} component retrieves the popularity scores from Wikipedia page view statistics, as shown in Table~\ref{tbl:case_study_candidate_answers}. Next, in the \textit{Debiasing} component, we adjust the plausibility scores using these popularity scores according to Equation~\ref{eq:debiasing}, with \(\alpha = 0.5\)\footnote{This hyperparameter is tunable; here we fix it at 0.5 for illustration.}.  

For example, for the candidate answer \textit{Ernst Hilgard}, we compute the debiased plausibility score as follows:

{\small
\begin{align}
    \label{eq:case_study_debiasing}
    DePls_4 &= Pls_4 \times \left( 1 - \alpha \times Pop_4 \right) \\
            &= 25 \times \left(1 - 0.5 \times 0.176 \right) \notag \\
            &= 25 \times 0.912 \notag \\
            &= 22.800 \notag
\end{align}
}

Table~\ref{tbl:case_study_candidate_answers} shows all resulting debiased plausibility scores.

\subsection{Scoring}
In the \textit{Entropy} component, we first normalize the debiased plausibility scores to a probability distribution using Equation~\ref{eq:pop_distribution}. Table~\ref{tbl:case_study_candidate_answers} reports these normalized values. For example, for the candidate answer \textit{Ernst Hilgard}, normalization proceeds as follows:

{\small
\begin{align}
    \label{eq:case_study_normalization}
    DePls_4^{\text{norm}} &= \frac{DePls_4}{\sum_i DePls_i} \\
                          &= \frac{22.800}{532.715} \notag \\
                          &= 0.043 \notag
\end{align}
}

We then compute the entropy~\cite{https://doi.org/10.1002/j.1538-7305.1948.tb01338.x} of the normalized debiased plausibility scores using Equation~\ref{eq:entropy}, obtaining \(H(q) = 3.9702\). Finally, the \textit{Normalization} component scales this entropy to produce the final difficulty score:

{\small
\begin{align}
    \label{eq:case_study_score}
    Diff_q &= \frac{H(q)}{\log_2 N} \\
           &= \frac{3.9702}{4.3219} \notag \\
           &= 0.91 \notag
\end{align}
}

\newpage

\section{Dataset Details}\label{apx:dataset_details}
\begin{table}[t]
\centering
\resizebox{\columnwidth}{!}{%
\begin{tabular}{@{}l@{\hspace{50pt}}ccc@{}}
\toprule
\textbf{Dataset}       & \textbf{Train} & \textbf{Validation} & \textbf{Test} \\ \midrule
\multicolumn{4}{c}{\textit{Simple}}                                          \\ \midrule
TriviaQA               & 78,785         & 8,837               & 11,313        \\
Natural Questions (NQ) & 79,168         & 8,757               & 3,610         \\ \midrule
\multicolumn{4}{c}{\textit{Complex}}                                      \\ \midrule
MuSiQue                & 19,938         & 2,417               & 2,459         \\
QASC                   & 8,134          & 926                 & 920           \\ \bottomrule
\end{tabular}%
}
\caption{Dataset statistics with standard train/validation/test splits.}
\label{tbl:dataset_details_statistics}
\end{table}

In this section, we provide detailed descriptions and key statistics for the datasets used in our experiments. These resources represent a diverse mix of simple and complex QA benchmarks, covering a wide range of domains and question types. Together, they establish a robust evaluation suite for analyzing the effectiveness and generalizability of \method method. The standard train/validation/test statistics for these datasets are summarized in Table~\ref{tbl:dataset_details_statistics}.

\paragraph{TriviaQA}
TriviaQA~\cite{joshi-etal-2017-triviaqa} is a large-scale QA dataset containing over 650{,}000 question–answer pairs collected from trivia websites. The questions are primarily factoid, requiring the identification of named entities or short factual responses. TriviaQA also includes approximately 95{,}000 questions that have been verified with supporting evidence from Wikipedia and the web.

\paragraph{Natural Questions (NQ)}
Natural Questions~\cite{kwiatkowski-etal-2019-natural} consists of around 300,000 real user queries submitted to Google Search, with annotations for both long and short answers over Wikipedia articles. Roughly 100,000 of these questions are categorized as factoid. In this work, we only focus on the subset of NQ questions included in AmbigQA~\cite{min-etal-2020-ambigqa}, which contains ambiguous questions paired with multiple valid factoid-style answers.

\paragraph{MuSiQue}
MuSiQue~\cite{trivedi-etal-2022-musique} (Multi-hop Structured Questions) is a multi-hop QA dataset with about 25,000 questions designed to require reasoning over multiple supporting facts. These questions are generally non-factoid and involve synthesizing evidence from different sources to reach the correct answer.

\paragraph{QASC}
QASC~\cite{Khot_Clark_Guerquin_Jansen_Sabharwal_2020} is a multiple-choice QA benchmark focused on elementary science, featuring 9,980 questions that require reasoning over simple scientific facts. Each question has eight candidate answer choices, and answering often demands combining multiple facts.

% \begin{table}[t]
% \centering
% \resizebox{\columnwidth}{!}{%
% \begin{tabular}{l@{\hspace{50pt}}cc}
% \toprule
% \textbf{Model} & \textbf{Category} & \textbf{Parameters} \\
% \midrule
% LLaMA 3.2 & Small & 3B \\
% Gemma 3 & Small & 4B \\
% \midrule
% Mistral & Medium & 7B \\
% Qwen 2.5 & Medium & 7B \\
% LLaMA 3.1 & Medium & 8B \\
% \midrule
% Mistral Small & Large & 24B \\
% Gemma 2 & Large & 27B \\
% \midrule
% LLaMA 3.1 & Very-Large & 70B \\
% Qwen 2.5 & Very-Large & 72B \\
% \midrule
% GPT-4 & Ultra-Large & not publicly disclosed \\
% \bottomrule
% \end{tabular}%
% }
% \caption{Summary of LLMs used in our experiments, including category and parameter size.}
% \label{tbl:model_details_statistics}
% \end{table}

\begin{table*}[t]
\centering
\resizebox{\textwidth}{!}{%
\begin{tabular}{l@{\hspace{50pt}}c@{\hspace{20pt}}c@{\hspace{20pt}}c@{\hspace{20pt}}c@{\hspace{20pt}}c@{\hspace{20pt}}c@{\hspace{20pt}}c@{\hspace{20pt}}c}
\toprule
\textbf{Model} & \textbf{Category} & \textbf{Params} & \textbf{Context} &
\textbf{Instr. Tuned} & \textbf{Weights Public} & \textbf{Provider} & \textbf{Year} \\
\midrule
LLaMA 3.2 & Small & 3B & 8K & Yes & Yes & Meta & 2024 \\
Gemma 3 & Small & 4B & 8K & Yes & Yes & Google & 2024 \\
\midrule
Mistral 7B & Medium & 7B & 8K & Yes & Yes & Mistral AI & 2023 \\
Qwen 2.5 & Medium & 7B & 32K & Yes & Yes & Alibaba & 2024 \\
LLaMA 3.1 & Medium & 8B & 8K & Yes & Yes & Meta & 2024 \\
\midrule
Mistral Small & Large & 24B & 32K & Yes & Yes & Mistral AI & 2024 \\
Gemma 2 & Large & 27B & 8K & Yes & Yes & Google & 2024 \\
\midrule
LLaMA 3.1 & Very-Large & 70B & 8K & Yes & Yes & Meta & 2024 \\
Qwen 2.5 & Very-Large & 72B & 32K & Yes & Yes & Alibaba & 2024 \\
\midrule
GPT-4 & Ultra-Large & N/A & N/A & Yes & No & OpenAI & 2023 \\
\bottomrule
\end{tabular}%
}
\caption{Characteristics of the LMs evaluated in our experiments, including architectural scale, context window size, instruction tuning, providers, and availability to support reproducibility.}
\label{tbl:model_details_statistics}
\end{table*}

\newpage

\section{Models Details}\label{apx:models_details}

\begin{table*}
\centering
\rowcolors{2}{white}{white} % reset default alternating colors
\resizebox{\textwidth}{!}{%
\begin{tabular}{c p{9.5cm} l c c}
\toprule
\textbf{\#} & \textbf{Question} & \textbf{Ground Truth} & \textbf{Difficulty Score} & \textbf{Group} \\
\midrule
\rowcolor{green!10}
1 & Which is the most powerful chess piece? & Queen & $0.2 \leq 0.69$ & Easy \\ 
\rowcolor{green!10}
2 & What product of photosynthesis, a carbohydrate occurring in the cells of plants, can be changed into glucose or dextrine? & Starch & $0.12 \leq 0.69$ & Easy \\ 
\rowcolor{green!10}
3 & Which state did frontiersman Davy Crockett represent in the US House of Representatives? & Tennessee & $0.15 \leq 0.69$ & Easy \\ 
\rowcolor{green!10}
4 & In which James Bond film does actress Jane Seymour play Solitaire? & Live and Let Die & $0.19 \leq 0.69$ & Easy \\ 
\rowcolor{green!10}
5 & `Sufferin' succotash' is a catchphrase of which cartoon cat? & Sylvester & $0.43 \leq 0.69$ & Easy \\ 
\rowcolor{red!10}
6 & What county is Moran located in the state where Konza Prairie Biological Station is located? & Kansas & $0.97 > 0.69$ & Hard \\ 
\rowcolor{red!10}
7 & Who is the spouse of the director of Jump for Glory? & Miriam Cooper & $0.99 > 0.69$ & Hard \\ 
\rowcolor{red!10}
8 & The artist adding backing vocals to You're So Vain attended what grammar school? & Mick Jagger, Dartford & $0.98 > 0.69$ & Hard \\ 
\rowcolor{red!10}
9 & What is the name of the airport in the city where WILM is licensed to broadcast? & Wilmington International Airport & $0.95 > 0.69$ & Hard \\ 
\rowcolor{red!10}
10 & What gun was used by Pollack's director in Westworld? & LeMat revolver & $0.99 > 0.69$ & Hard \\
\bottomrule
\end{tabular}%
}
\caption{List of questions sampled from the TriviaQA and MuSiQue datasets, along with their ground truths, difficulty scores, and group labels. The \textcolor{green}{green} rows indicate easy questions, where the difficulty score is below the threshold, while the \textcolor{red}{red} rows indicate hard questions, where the difficulty score exceeds the threshold.}
\label{tbl:cohen_d_questions}
\end{table*}

In this section, we provide more detailed information about the LMs used in our experiments. We categorize these LMs into five groups: \textit{Small LMs}, \textit{Medium LMs}, \textit{Large LMs}, \textit{Very-Large LMs}, and \textit{Ultra-Large LMs}. Table~\ref{tbl:model_details_statistics} summarizes each LM along with its category and parameter size.

\paragraph{\textbf{Small LMs}}
\textit{LLaMA 3.2 3B}~\cite{grattafiori2024llama3herdmodels} is a compact yet capable model optimized for fast inference and moderate computational efficiency.
\textit{Gemma 3 4B}~\cite{gemmateam2025gemma3technicalreport} is a lightweight transformer model designed for efficient language processing while maintaining competitive performance on standard NLP benchmarks.

\paragraph{\textbf{Medium LMs}}
\textit{Mistral 7B}~\cite{jiang2023mistral7b} is a high-performance, decoder-only model known for balanced efficiency and effectiveness in generative tasks.
\textit{Qwen 2.5 7B}~\cite{qwen2025qwen25technicalreport} is a 7-billion parameter model trained for robust text generation, excelling in multilingual and knowledge-intensive tasks.
\textit{LLaMA 3.1 8B}~\cite{grattafiori2024llama3herdmodels} is an improved version of LLaMA designed for enhanced reasoning and generalization in language tasks.

\paragraph{\textbf{Large LMs}}
\textit{Mistral 24B}~\cite{jiang2023mistral7b} is a larger variant designed for high-capacity reasoning and advanced generative performance, with a strong ability to follow complex instructions.
\textit{Gemma 2 27B}~\cite{gemmateam2024gemma2improvingopen} is a mid-scale model offering an excellent trade-off between capacity and computational cost, well-suited for general-purpose language tasks.

\paragraph{\textbf{Very-Large LMs}}
\textit{LLaMA 3.1 70B}~\cite{grattafiori2024llama3herdmodels} is a large-scale transformer model optimized for complex reasoning, long-form generation, and open-domain QA.
\textit{Qwen 2.5 72B}~\cite{qwen2025qwen25technicalreport} is a state-of-the-art multilingual model with extensive parameter tuning for diverse NLP applications.

\paragraph{\textbf{Ultra-Large LMs}}
\textit{GPT-4}~\cite{openai2024gpt4technicalreport} is a highly sophisticated language model with superior reasoning, comprehension, and problem-solving abilities across multiple domains, representing a leading frontier in LLM research.

\newpage

\section{Metrics Details}\label{apx:metrics_details}
\begin{figure}
    \centering
    \begin{tcolorbox}[
        enhanced,
        width=\columnwidth,
        colback=black!2,
        colframe=black!60,
        boxrule=0.5pt,
        arc=2mm,
        left=1.5mm,
        right=1.5mm,
        top=1.2mm,
        bottom=1.2mm,
        title=\textbf{\raisebox{0pt}[2ex][1ex]{Evaluating the Generated Answer using LLM}},
        fonttitle=\bfseries\scriptsize,
        colbacktitle=black!20,
        coltitle=black,
        boxed title style={
            colframe=black!60,
            colback=black!10,
            boxrule=0.5pt,
            arc=2mm,
            valign=center
        }
    ]
    \scriptsize
    Question: $<$question$>$ \\
    Answer: $<$ground\_truth$>$ \\
    Candidate: $<$candidate$>$ \\
    Is the candidate correct? Choose between "Yes" or "No"
    \end{tcolorbox}
    
    \caption{The placeholder \texttt{<question>} represents the question, \texttt{<ground\_truth>} indicates the correct answer, and \texttt{<candidate>} shows the answer generated by different LLMs.}
    
    \label{fig:gpt_eval_prompt}
\end{figure}

In this section, we provide detailed descriptions of the evaluation metrics used throughout the paper, including both \texttt{GPT-Eval} and Cohen's $d$.

\subsection{GPT-Eval}\label{apx:gpt-eval}

In this section, we describe the motivation of using \texttt{GPT-Eval} instead of token-based metrics and then how it works.

In the era of LLMs, traditional metrics such as Exact Match (EM) or token-based \textit{Contains} measures are often insufficient, since LLMs often generate correct answers in varied phrasings. For example, for the question \textit{What is the term for a trained professional responsible for diagnosing and treating illnesses?}, answers like \textit{Doctor}, \textit{Medic}, or \textit{Practitioner} are all acceptable, but EM or token-overlap metrics would fail to recognize their equivalence. LLM-based evaluators can handle such semantic equivalence, motivating our choice of \texttt{GPT-Eval} over traditional metrics.

In the \texttt{GPT-Eval} framework, we prompt an LLM using the template shown in Figure~\ref{fig:gpt_eval_prompt}, asking it to determine whether a candidate answer is correct for a given question, with access to the corresponding ground-truth answer. The model outputs a binary judgment: \textit{Yes} or \textit{No}.

\begin{table}
% \small
\centering
\resizebox{0.9\columnwidth}{!}{%
\begin{tabular}{l@{\hspace{100pt}}l}
\toprule
\textbf{Cohen's $d$} & \textbf{Interpretation} \\
\midrule
$d < 0.00$ & Very Weak \\
$0.00 \leq d < 0.20$ & Weak \\
$0.20 \leq d < 0.50$ & Modest \\
$0.50 \leq d < 0.80$ & Moderate \\
$0.80 \leq d < 1.20$ & Strong \\
$1.20 \leq d$ & Very Strong \\
\bottomrule
\end{tabular}%
}
\caption{Interpretation of Cohen's $d$ thresholds}
\label{tbl:cohens_d_interpretation}
\end{table}

\begin{table*}
\centering
\resizebox{\textwidth}{!}{%
\begin{tabular}{@{}l |p{2.5cm} p{3cm} p{4cm} p{6cm} p{5cm} |c@{}}
\toprule
\multicolumn{7}{c}{\textit{Easy Questions}} \\
\midrule
\textbf{LLM} & \textbf{Q1} & \textbf{Q2} & \textbf{Q3} & \textbf{Q4} & \textbf{Q5} & \textbf{Acc} \\
\midrule
LLaMA 3.2 3b & \cellcolor{green!20}Queen & \cellcolor{green!20}Starch & \cellcolor{green!20}Tennessee & \cellcolor{green!20}Live and Let Die & Popeye & 0.8 \\
LLaMA 3.1 8b & \cellcolor{green!20}Queen & \cellcolor{green!20}Starch & \cellcolor{green!20}Tennessee & \cellcolor{green!20}Live and Let Die & \cellcolor{green!20}Sylvester the Cat & 1.0 \\
Mistral 7B & KNIGHT & \cellcolor{green!20}STARCH & \cellcolor{green!20}Tennessee & \cellcolor{green!20}LIVE AND LET DIE & Garfield & 0.6 \\
Qwen 2.5 7b & \cellcolor{green!20}Queen & \cellcolor{green!20}Starch & \cellcolor{green!20}Tennessee & Licence to Kill & Tom & 0.8 \\
LLaMA 3.1 70b & \cellcolor{green!20}Queen & \cellcolor{green!20}Starch & \cellcolor{green!20}Tennessee & \cellcolor{green!20}Live and Let Die & \cellcolor{green!20}Sylvester & 1.0 \\
Qwen 2.5 72b & \cellcolor{green!20}Queen & \cellcolor{green!20}Starch & \cellcolor{green!20}Tennessee & \cellcolor{green!20}Live and Let Die & \cellcolor{green!20}Sylvester & 1.0 \\
Gemma 2.0 27b & \cellcolor{green!20}Queen & \cellcolor{green!20}Starch & \cellcolor{green!20}Tennessee & \cellcolor{green!20}Live and Let Die & \cellcolor{green!20}Sylvester & 1.0 \\
Mistral 24b & \cellcolor{green!20}Queen & \cellcolor{green!20}Starch & \cellcolor{green!20}Tennessee & \cellcolor{green!20}Live and Let Die & \cellcolor{green!20}Sylvester & 1.0 \\
Gemma 3.0 4b & \cellcolor{green!20}Queen & \cellcolor{green!20}Starch & Texas & A View to a Kill & Tom Cat & 0.4 \\
GPT 4 & \cellcolor{green!20}Queen & \cellcolor{green!20}Starch & \cellcolor{green!20}Tennessee & \cellcolor{green!20}Live and Let Die & \cellcolor{green!20}Sylvester & 1.0 \\
\midrule
\multicolumn{7}{c}{\textit{Hard Questions}} \\
\midrule
\textbf{LLM} & \textbf{Q6} & \textbf{Q7} & \textbf{Q8} & \textbf{Q9} & \textbf{Q10} & \textbf{Acc} \\
\midrule
LLaMA 3.2 3b & \cellcolor{green!20}Kansas & Linda Gray & Charterhouse & \cellcolor{green!20}Wilmington & Smith \& Wesson & 0.4 \\
LLaMA 3.1 8b & Wabaunsee & Michael Jordan & Bedales & \cellcolor{green!20}Wilmington & Colt Python & 0.2 \\
Mistral 7B & \cellcolor{green!20}Kansas & NO ANSWER & Beverly Hills High School & NO ANSWER & GUNSLINGER & 0.2 \\
Qwen 2.5 7b & Douglas & Lynne Ramsay & Westminster School & Salisbury Airport & Smith \& Wesson Model 29 & 0.0 \\
LLaMA 3.1 70b & Geary & Michael Powell & \cellcolor{green!20}Mick Jagger & \cellcolor{green!20}Wilmington International Airport & Colt Peacemaker & 0.4 \\
Qwen 2.5 72b & Riley & Lena Waithe & Radley College & Newark Liberty International Airport & Colt Python & 0.0 \\
Gemma 2.0 27b & Riley & Jennifer Lopez & Miss Porter’s School & \cellcolor{green!20}Wilmington International Airport & Mauser C96 & 0.2 \\
Mistral 24b & Riley & Sally Phillips & Radley College & New Castle & Colt Single Action Army & 0.0 \\
Gemma 3.0 4b & Finney & Lara & St. Paul’s & Philadelphia International & \cellcolor{green!20}Revolver & 0.2 \\
GPT 4 & Riley & Tom Hanks & \cellcolor{green!20}Mick Jagger & \cellcolor{green!20}Wilmington International Airport & Colt Single Action Army & 0.4 \\
\bottomrule
\end{tabular}%
}
\caption{LLM responses on both Easy and Hard questions with their accuracy scores. \textcolor{green}{Green} colored cells indicate correct answers.}
\label{tbl:cohens_d_llm_evaluated_questions}
\end{table*}

\subsection{Cohen's $d$}\label{apx:cohens_d}

In this section, we present an example to clarify the use of Cohen’s $d$~\cite{Cohen2013-zu} and illustrate our evaluation procedure. We follow the steps described in Section~\ref{sss:cohen_d} to explain this example in detail. Specifically, we sampled $10$ questions from the TriviaQA~\cite{joshi-etal-2017-triviaqa} and MuSiQue~\cite{trivedi-etal-2022-musique} datasets, with $\alpha=0.5$. Table~\ref{tbl:cohen_d_questions} shows the sampled questions along with their gold answers, computed difficulty scores, and assigned groups. The median of these difficulty scores is $0.69$, so we categorize questions with a difficulty score below $0.69$ as \textit{Easy} and those above $0.69$ as \textit{Hard}. As shown in the \textit{Group} column of Table~\ref{tbl:cohen_d_questions}, five questions fall into the \textit{Easy} group and five into the \textit{Hard} group.

Next, we prompted the ten LLMs described in Section~\ref{ss:models} to answer each question in both groups using the prompt illustrated in Figure~\ref{fig:qa_prompt}. Table~\ref{tbl:cohens_d_llm_evaluated_questions} presents the answers generated by each LLM for both the \textit{Easy} and \textit{Hard} groups. As shown in this table, some generated answers are not lexically identical to the ground truth but are still considered correct, thanks to the use of \texttt{GPT-Eval}~\cite{kamalloo-etal-2023-evaluating} for semantic evaluation. For instance, for the question \textit{‘Sufferin' succotash’ is a catchphrase of which cartoon cat?}, LLaMA 3.1 8B produces \textit{Sylvester the Cat}, which is not lexically identical to the ground truth \textit{Sylvester}, but is semantically equivalent and marked as correct by \texttt{GPT-Eval}. The \textit{Acc} column shows the accuracy of each LLM as a QA system, calculated using Equation~\ref{eq:accuracy}. As expected, the accuracy on easy questions is higher than on hard questions for each LLM, confirming that harder questions are indeed more challenging in practice.

\begin{figure}[h!]
    \centering
    \begin{tcolorbox}[
        enhanced,
        width=\columnwidth,
        colback=black!2,
        colframe=black!60,
        boxrule=0.5pt,
        arc=2mm,
        left=1.5mm,
        right=1.5mm,
        top=1.2mm,
        bottom=1.2mm,
        title=\textbf{\raisebox{0pt}[2ex][1ex]{Answer Generation}},
        fonttitle=\bfseries\scriptsize,
        colbacktitle=black!20,
        coltitle=black,
        boxed title style={
            colframe=black!60,
            colback=black!10,
            boxrule=0.5pt,
            arc=2mm,
            valign=center
        }
    ]
    \scriptsize
    \textbf{System:}    \newline
    You are an assistant that answers questions. You just answer questions with exact and short answers. You do not use sentences as the response.
    \newline
    \newline
    \textbf{Shot (1):}    \newline
    Answer the question under conditions: 1) Answer should not be sentences. It should be some words. 2) Do not generate "sorry" or "I cannot ..." sentences. 3) Do not generate explanations, reasoning, or full sentences—only provide the exact answer.
    
    Question: Who won the Nobel Peace Prize in 2009?
    
    Answer: Barack Obama
    \newline
    \newline
    
    \textbf{Shot (2):}    \newline
    Answer the question under conditions: 1) Answer should not be sentences. It should be some words. 2) Do not generate "sorry" or "I cannot ..." sentences. 3) Do not generate explanations, reasoning, or full sentences—only provide the exact answer.
    
    Question: Edouard Daladier became Prime Minister of which country in 1933?
    
    Answer: France
    \newline

    \textbf{Shot (3):}  ...   \newline
    \textbf{Shot (4):}  ...  \newline
    \textbf{Shot (5):}  ...  \newline
    
    \textbf{User:}    \newline
    Answer the question under conditions: 1) Answer should not be sentences. It should be some words. 2) Do not generate "sorry" or "I cannot ..." sentences. 3) Do not generate explanations, reasoning, or full sentences—only provide the exact answer.
    
    Question: Who played the character Pink in Pink Floyd: The Wall?
    
    Answer: $<$ANSWER$>$
    \end{tcolorbox}
    
    \caption{This prompt uses five-shot examples to guide the LLM in answering questions effectively. The placeholder \texttt{<ANSWER>} represents the answer generated by the LLM for the user’s question.}
    \label{fig:qa_prompt}
\end{figure}

We then computed the mean and standard deviation of the accuracy for each group separately. For the easy group, the mean accuracy is $0.86$ with a standard deviation of $0.2$; for the hard group, the mean is $0.2$ with a standard deviation of $0.154$. Finally, we calculate Cohen’s $d$ using Equation~\ref{eq:cohens_d}:
\begin{align}
    d &= \frac{0.86 - 0.2}{ \sqrt{ \frac{ 0.2^2 + 0.154^2 }{2} } } \\
      &= 3.697 \notag
\end{align}
According to Table~\ref{tbl:cohens_d_interpretation}, which shows the interpretation of Cohen’s $d$ values, $d=3.697$ is considered \textbf{very strong}, indicating that our method computes the difficulty score reliably and effectively separates questions into \textit{Easy} and \textit{Hard} groups.

\newpage

\section{Additional Experiments}\label{apx:experiments}
\begin{figure}
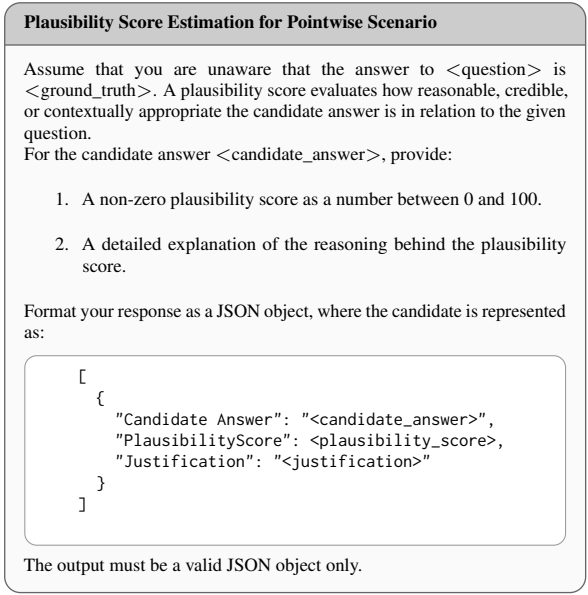

    \centering
    \begin{tcolorbox}[
        enhanced,
        width=\columnwidth,
        colback=black!2,
        colframe=black!60,
        boxrule=0.5pt,
        arc=2mm,
        left=1.5mm,
        right=1.5mm,
        top=1.2mm,
        bottom=1.2mm,
        title=\textbf{\raisebox{0pt}[2ex][1ex]{Plausibility Score Estimation for Pointwise Scenario}},
        fonttitle=\bfseries\scriptsize,
        colbacktitle=black!20,
        coltitle=black,
        boxed title style={
            colframe=black!60,
            colback=black!10,
            boxrule=0.5pt,
            arc=2mm,
            valign=center
        }
    ]
    \scriptsize
    Assume that you are unaware that the answer to $<$question$>$ is $<$ground\_truth$>$. A plausibility score evaluates how reasonable, credible, or contextually appropriate the candidate answer is in relation to the given question.    
    
    For the candidate answer $<$candidate\_answer$>$, provide:
    \begin{enumerate}
        \item A non-zero plausibility score as a number between 0 and 100.
        \item A detailed explanation of the reasoning behind the plausibility score.
    \end{enumerate}
    
    Format your response as a JSON object, where the candidate is represented as:
    \begin{tcolorbox}[
        colback=white,
        colframe=black!40,
        boxrule=0.4pt,
        arc=1.5mm,
        left=1mm,
        right=1mm,
        top=0.8mm,
        bottom=0.8mm
    ]
    \ttfamily\scriptsize
    \begin{verbatim}
    [
      {
        "Candidate Answer": "<candidate_answer>",
        "PlausibilityScore": <plausibility_score>,
        "Justification": "<justification>"
      }
    ]
    \end{verbatim}
    \end{tcolorbox}
    
    The output must be a valid JSON object only.
    \end{tcolorbox}
    
    \caption{The placeholder \texttt{<question>} represents the given question, while \texttt{<ground\_truth>} denotes its correct answer. Candidate answer is represented by \texttt{<candidate\_answer>}, along with a plausibility score (\texttt{<plausibility\_score>}) and a justification (\texttt{<justification>}) explaining both the answer choice and the reasoning behind its assigned score.}
    \label{fig:pointwise_prompt}
\end{figure}

In this section, we present additional experiments and supplementary materials that support and extend our main findings.

\subsection{Answer Plausibility Estimation Methods}\label{apx:answer_plausibility_estimation_methods}

To estimate the plausibility scores for candidate answers, we investigate three different methods: \textit{Pointwise}, \textit{Pairwise}, and \textit{Listwise}.  
For the \textit{Pointwise} scenario, we use the candidate answers generated using the prompt shown in Figure~\ref{fig:listwise_prompt} and then prompt the LLM separately for each candidate answer to estimate its plausibility score, using the prompt shown in Figure~\ref{fig:pointwise_prompt}.  
For the \textit{Pairwise} scenario, we estimate plausibility scores by comparing pairs of candidate answers to determine which is more likely correct for a given question. Formally, given the set of candidate answers $\mathcal{C}_q$ for a question $q$, we construct the set of pairs as $\{(x_1, x_2) \mid x_1, x_2 \in \mathcal{C}_q, \ x_1 \neq x_2 \}$. For each pair $(x_1, x_2)$, we prompt the LLM using the prompt shown in Figure~\ref{fig:pairwise_prompt} in Appendix~\ref{apx:pairwise_estimation_prompt} to compare them. After collecting preferences across all pairs for a given question, we apply the Bradley-Terry model~\cite{19ff28b9-64f9-3656-ba40-08326a05748e} to convert the pairwise comparisons into final plausibility scores.
For the \textit{Listwise} scenario, we use the prompt shown in Figure~\ref{fig:listwise_prompt} to directly generate candidate answers along with their plausibility scores.

To compare the runtime complexity of each scenario, we consider each prompt as having complexity $\mathcal{O}(1)$. Under this assumption, the runtime complexity for the \textit{Listwise} scenario is $\mathcal{O}(1)$ as it requires only a single prompt, while the \textit{Pointwise} and \textit{Pairwise} scenarios have complexities of $\mathcal{O}(n)$ and $\mathcal{O}(n^2)$, respectively. This indicates that \textit{Listwise} is also more efficient than the other scenarios in terms of runtime complexity.

Regarding the output length, the \textit{Pointwise} and \textit{Listwise} scenarios have approximately the same overall output length, since they ultimately produce the same candidate answer data, but \textit{Pointwise} does so across $n$ separate prompts, whereas \textit{Listwise} produces it in a single prompt. In contrast, the output length for the \textit{Pairwise} scenario is significantly larger because each pairwise comparison includes a separate justification. Table~\ref{tbl:scenarios_comparison} summarizes this comparison between the scenarios.

In Appendix~\ref{apx:pointwise_estimation_prompt} and~\ref{apx:pairwise_estimation_prompt}, we provide examples illustrating why 
% these methods
\textit{Pointwise} and \textit{Pairwise} methods may not estimate plausibility scores effectively.

\begin{table}
% \small
\centering
\resizebox{\columnwidth}{!}{%
\begin{tabular}{l@{\hspace{50pt}}cc}
\toprule
\textbf{Scenario} & \textbf{Runtime Complexity} & \textbf{Avg. Length} \\
\midrule
Pointwise & $\mathcal{O}(n)$ & 60.21 \\
Pairwise  & $\mathcal{O}(n^2)$ & 295.99 \\
Listwise  & $\mathcal{O}(1)$ & 55.91 \\
\bottomrule
\end{tabular}%
}
\caption{Comparison of runtime complexity and average output length (measured in words) across different scenarios for each question.}
\label{tbl:scenarios_comparison}
\end{table}

\subsection{Pointwise Estimation}\label{apx:pointwise_estimation_prompt}
\begin{figure*}[t]
	\centering
	\includegraphics[width=\textwidth]{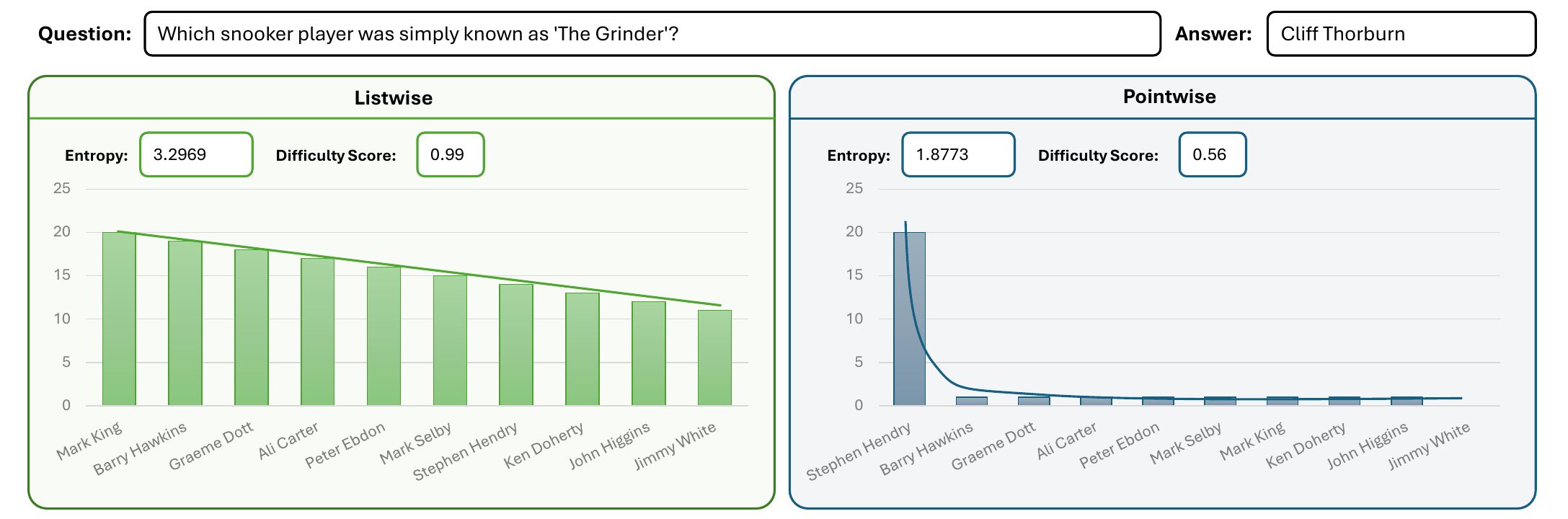}
    \caption{An example from TriviaQA of the pointwise scenario showing that most plausibility scores for the candidate answers are one, while in the listwise scenario, the candidate answers receive more diverse scores.}
    \label{fig:pointwise_example}
\end{figure*}
\begin{figure}[t]
    \centering
    \begin{tcolorbox}[
        enhanced,
        width=\columnwidth,
        colback=black!2,
        colframe=black!60,
        boxrule=0.5pt,
        arc=2mm,
        left=1.5mm,
        right=1.5mm,
        top=1.2mm,
        bottom=1.2mm,
        title=\textbf{\raisebox{0pt}[2ex][1ex]{Plausibility Comparison in the Pairwise Scenario}},
        fonttitle=\bfseries\scriptsize,
        colbacktitle=black!20,
        coltitle=black,
        boxed title style={
            colframe=black!60,
            colback=black!10,
            boxrule=0.5pt,
            arc=2mm,
            valign=center
        }
    ]
    \scriptsize
    Choose which of the two candidate answers is more likely to be correct based on the given question. A 'better candidate answer' is the one with a higher probability of being correct. Assume you do not know that the correct answer is $<$ground\_truth$>$.
    \\
    \\
    
    Question: $<$question$>$  \\
    Candidate Answer 1: $<$candidate\_1$>$ \\
    Candidate Answer 2: $<$candidate\_2$>$  
    \\
    \\
    
    Which is the better candidate answer? Respond with only "1" if \texttt{Candidate Answer 1} is better, or "2" if \texttt{Candidate Answer 2} is better. Provide a justification, and your final response must be a single character: "1" or "2".
    \end{tcolorbox}
    
    \caption{The placeholder \texttt{<question>} represents the given question, while \texttt{<ground\_truth>} denotes its correct answer. Two candidate answers, \texttt{<candidate\_1>} and \texttt{<candidate\_2>}, are compared, and the LLM is asked to select which one is more likely to be correct, along with a justification.}

    \label{fig:pairwise_prompt}
\end{figure}
\begin{figure}[t]
	\centering
	\includegraphics[width=\columnwidth]{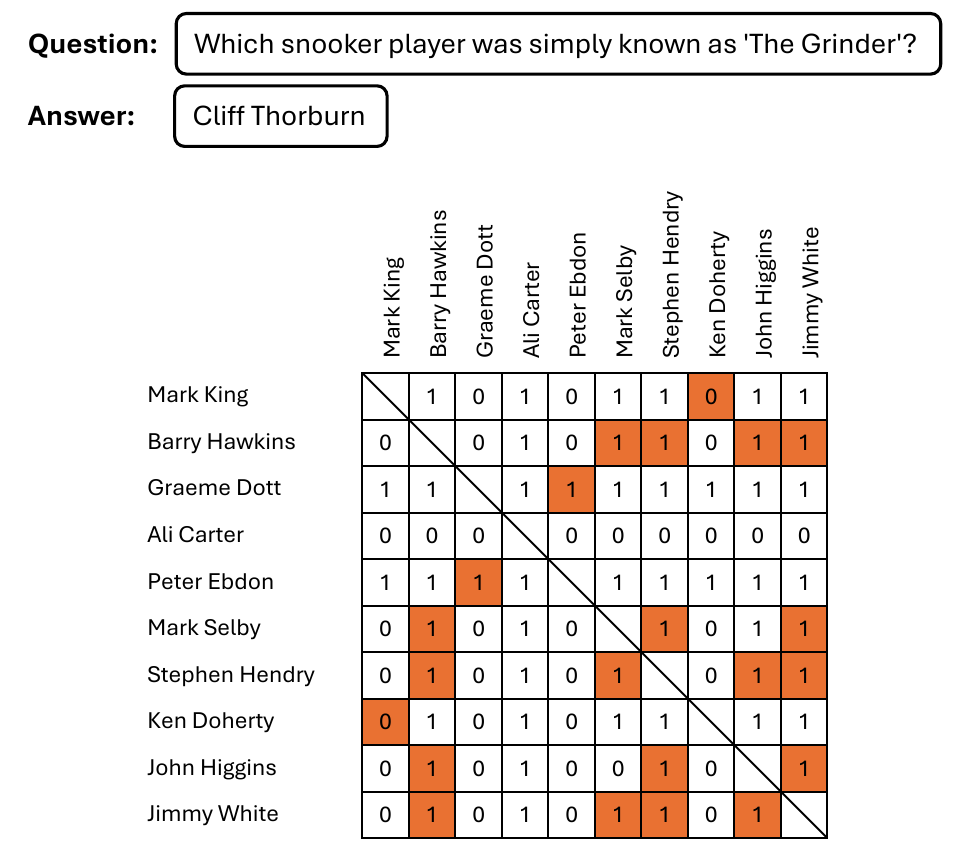}
    \caption{An example from TriviaQA of the pairwise scenario showing the pairwise comparison matrix. A value of $1$ indicates that the candidate answer in the row is preferred over the candidate answer in the column, while a value of $0$ indicates the opposite. \textcolor{red}{Red} cells highlight inconsistencies in the pairwise preferences.}

    \label{fig:pairwise_example}
\end{figure}

To estimate plausibility scores in the pointwise scenario, we use the prompt shown in Figure~\ref{fig:pointwise_prompt}. In this prompt, the LLM is instructed to estimate a plausibility score for a single candidate answer without knowledge of the other candidate answers. As a result, the LLM evaluates each candidate independently.\looseness=-1

Figure~\ref{fig:pointwise_example} presents an example illustrating the plausibility scores generated for the question \textit{Which snooker player was simply known as 'The Grinder'?} from TriviaQA using the \textit{Pointwise} and \textit{Listwise} scenarios. As shown, most candidate answers in the pointwise scenario receive a plausibility score of one. This pattern appears frequently, indicating that the pointwise scenario struggles to estimate plausibility scores effectively and tends to assign a score of one to most candidates. In contrast, the listwise scenario produces a more diverse range of plausibility scores. Additionally, this figure demonstrates how the limitations of the pointwise scenario can affect the entropy and the final difficulty score. For the same question, the difficulty score under the pointwise scenario is $0.56$, whereas under the listwise scenario it is $0.99$. This discrepancy further motivates our preference for the listwise scenario beyond its computational efficiency.\looseness=-1

\begin{figure*}
	\centering
	\includegraphics[width=\linewidth]{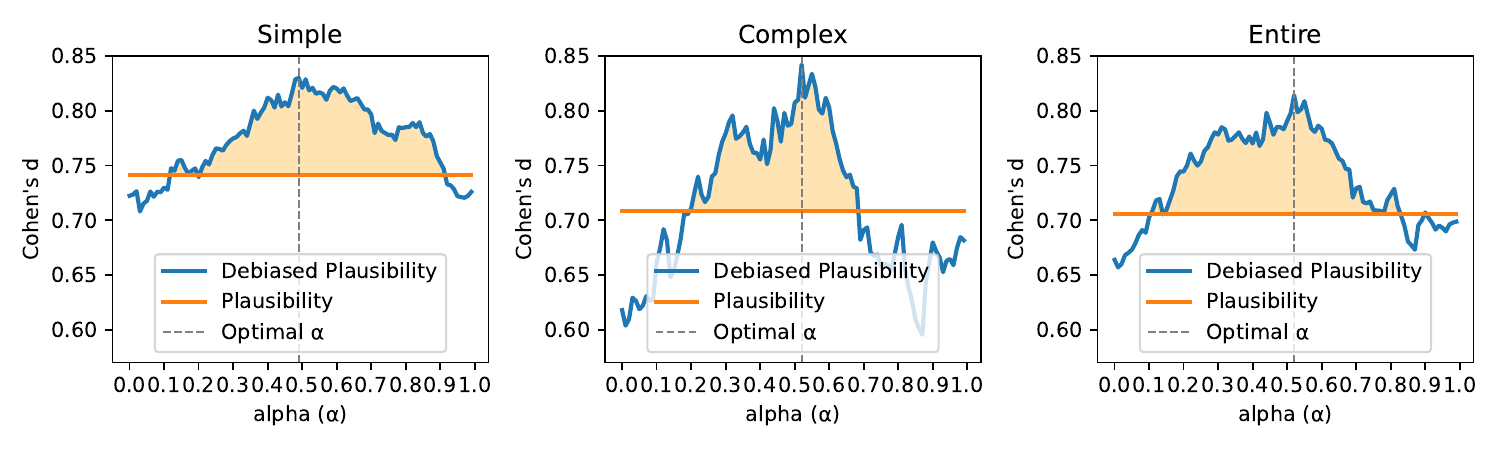}
	\caption{$\alpha$ robustness across different values from 0 to 1 on various question categories, including Simple, Complex, and Entire question types.}
	\label{fig:robustness}
\end{figure*}

\subsection{Pairwise Estimation}\label{apx:pairwise_estimation_prompt}

To estimate plausibility scores in the pairwise scenario, we use the prompt shown in Figure~\ref{fig:pairwise_prompt}, which presents a pair of candidate answers and asks the LLM to identify the more likely correct one.

Figure~\ref{fig:pairwise_example} presents an example illustrating the preferences generated for the question \textit{Which snooker player was simply known as 'The Grinder'?} from TriviaQA using the \textit{Pairwise} scenario. In this figure, a value of $1$ indicates that the candidate answer in the row is preferred over the candidate answer in the column, while $0$ indicates the opposite. As shown, there are many inconsistencies in the pairwise comparison matrix, highlighted in red. These inconsistencies may arise from the hallucination problem in LLMs. Furthermore, since the pairwise scenario requires a large number of prompts per question, the likelihood of hallucinations increases, which can directly affect the final pairwise scores by introducing more inconsistencies into the matrix. This discrepancy further motivates our preference for the listwise scenario beyond its computational efficiency.

\subsection{Generalization}\label{apx:generalization}
\begin{table}
\centering
\resizebox{\columnwidth}{!}{%
\begin{tabular}{lccc}
\toprule
\textbf{Question Type} & \textbf{Optimal} $\alpha$ & \textbf{Optimal \# of Cans} & $d$ \\
\midrule
Simple            & 0.49 & 7 & 0.8299 \\
Complex        & 0.52 & 8 & 0.8423 \\
Entire             & 0.52 & 8 & 0.8142 \\
\bottomrule
\end{tabular}%
}
\caption{Summary of optimal $\alpha$, optimal number of candidates, and Cohen's $d$ for simple, complex, and entire cases.}
\label{tbl:generalization}
\end{table}

In this experiment, we measure the most optimal number of candidate answers and the best $\alpha$ by grouping questions according to broader categories, rather than by dataset. Specifically, we compute these optimal values for question categories such as \textit{Simple} (TriviaQA and NQ datasets) and \textit{Complex} (QASC and MuSiQue datasets), which can provide more practical guidance for reproducibility and for users selecting settings based on their question types. Additionally, we determine the optimal values across the entire dataset (combining simple and complex questions) to provide general-purpose recommendations that can be applied regardless of question type.

Table~\ref{tbl:generalization} shows that the optimal $\alpha$ values for all cases are very close to each other and cluster around $0.5$, suggesting that the method is not dependent on question type. Similarly, the optimal number of candidate answers is consistent across cases and is significantly smaller than the initial maximum of $20$, indicating that generating a high number of candidates is not necessary to achieve strong results. Finally, the reported Cohen’s $d$ values show that, with these configurations, question difficulty separation remains very strong, highlighting the independence and robustness of the \method method across different question types.

\subsection{$\alpha$ Robustness}\label{apx:alpha_robustness}

In this experiment, we demonstrate that the superiority of \textit{Debiased-Plausibility} scenario is not limited to a specific configuration, particularly the parameter $\alpha$. We categorize the datasets and questions into three main groups based on question type: \textit{Simple} (TriviaQA and NQ datasets), \textit{Complex} (QASC and MuSiQue datasets), and \textit{Entire} (all datasets combined). We then compute the Cohen’s $d$ values for various $\alpha$ values ranging from 0 to 1, and compare them with the best configuration of the \textit{Plausibility} scenario. 

Figure~\ref{fig:robustness} illustrates the Cohen’s $d$ values for these different scenarios. The results indicate that, across all cases and for most $\alpha$ values, the Cohen’s $d$ for \textit{Debiased-Plausibility} exceeds that of the \textit{Plausibility} scenario. This suggests that the \method approach does not rely on a narrow configuration to achieve strong results, but rather outperforms the second-best method across a broad range of settings.

\begin{figure*}
	\centering
	\includegraphics[width=0.9\linewidth]{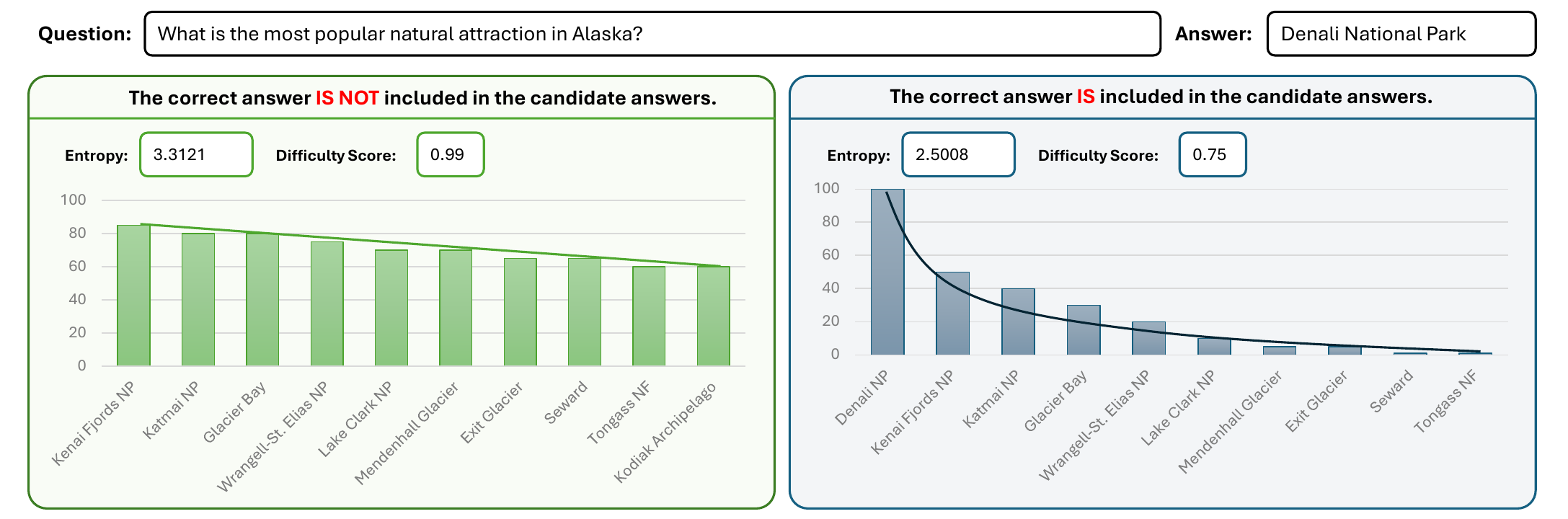}
	\caption{An example from the TriviaQA dataset illustrating how including the correct answer among the candidate answers can affect plausibility scores. The \textcolor{green}{green} box shows the case where the correct answer is excluded, while the \textcolor{blue}{blue} box shows how including it can influence the plausibility distribution.}
	\label{fig:gold_answer_inclusion}
\end{figure*}
\begin{figure}[h!]
    \centering
    \begin{tcolorbox}[
        enhanced,
        width=\columnwidth,
        colback=black!2,
        colframe=black!60,
        boxrule=0.5pt,
        arc=2mm,
        left=1.5mm,
        right=1.5mm,
        top=1.2mm,
        bottom=1.2mm,
        title=\textbf{\raisebox{0pt}[2ex][1ex]{Candidate Answer Generation with Excluding the Correct Answer}},
        fonttitle=\bfseries\scriptsize,
        colbacktitle=black!20,
        coltitle=black,
        boxed title style={
            colframe=black!60,
            colback=black!10,
            boxrule=0.5pt,
            arc=2mm,
            valign=center
        }
    ]
    \scriptsize
    Generate a list of 20 unique candidate answers. A plausibility score evaluates how reasonable, credible, or contextually appropriate each candidate answer is in relation to the given question. For each candidate, provide:
    \begin{enumerate}
        \item A non-zero plausibility score as a number between 0 and 100.
        \item A detailed explanation of the reasoning behind the plausibility score.
    \end{enumerate}
    
    Format your response as a JSON list, where each candidate is represented as:
    \begin{tcolorbox}[
        colback=white,
        colframe=black!40,
        boxrule=0.4pt,
        arc=1.5mm,
        left=1mm,
        right=1mm,
        top=0.8mm,
        bottom=0.8mm
    ]
    \begin{verbatim}
    [
      {
        "Candidate Answer": "<candidate_answer>",
        "PlausibilityScore": <plausibility_score>,
        "Justification": "<justification>"
      }
    ]
    \end{verbatim}
    \end{tcolorbox}
    
    The output must be a valid JSON list only.
    \end{tcolorbox}
    \caption{The placeholder \texttt{<question>} represents the given question. Each candidate answer is denoted by \texttt{<candidate\_answer>}, accompanied by an initial plausibility score (\texttt{<plausibility\_score>}) and a justification (\texttt{<justification>}) explaining both the answer choice and the rationale behind its plausibility score.}
    \label{fig:prompt_golden_answer_inclusion}
\end{figure}

\newpage

\section{Gold Answer Inclusion}\label{apx:gold_answer_inclusion}

In this section, we show the impact of including the correct answer among candidate answers using a motivating example. We prompt the LLaMA 3.3~\cite{grattafiori2024llama3herdmodels} model to generate 10 candidate answers\footnote{We use 10 candidates for clearer visualization and analysis compared to all 20.} to the question \textit{What is the most popular attraction in Alaska?}, whose answer is \textit{Denali National Park}, under two scenarios:

\begin{enumerate}
\item We use our standard prompt for generating candidate answers and plausibility scores, as shown in Figure~\ref{fig:listwise_prompt}. This prompt explicitly instructs the LLM not to include the correct answer as a candidate, while still providing it in the prompt to guide plausibility estimates.
\item We modify the prompt to remove the instruction excluding the correct answer, allowing the LLM to rely on its knowledge to identify and potentially include the correct answer among the candidates. The modified prompt used for this setting is shown in Figure~\ref{fig:prompt_golden_answer_inclusion}.
\end{enumerate}

To ensure consistency and reproducibility, we set the temperature to zero in both scenarios, so the same candidate answers could be generated, with the only difference being whether the LLM was allowed to include the correct answer. Figure~\ref{fig:gold_answer_inclusion} shows the plausibility scores for both conditions.

As shown, including the correct answer in scenario 2 distorts the plausibility distribution: the correct answer receives a plausibility score of $100$, while scores for other candidate answers drop significantly compared to scenario 1. This happens because the LLM assigns plausibility scores to incorrect candidates relative to the correct answer, leading to very low plausibility for all candidates. In contrast, scenario 1, where the model is explicitly instructed to ignore the correct answer as a candidate, allows the LLM to assign more balanced plausibility scores among the candidate answers.

\begin{table*}[t]
\centering
\small
\begin{tabular}{p{3.2cm} p{3.6cm} p{4.2cm} p{3.6cm}}
\toprule
\textbf{Pattern} & \textbf{Observed Behavior} & \textbf{Primary Cause} & \textbf{Implication} \\
\midrule
Entropy-overestimated difficulty &
High entropy; QA models answer correctly &
Dense set of semantically related plausible answers (e.g., list-based factual questions) &
Entropy reflects answer-space richness rather than model uncertainty \\

Entropy-underestimated difficulty &
Low entropy; QA models fail exact-match evaluation &
Ambiguous or underspecified gold annotations (e.g., nationality, definitions) &
Low entropy does not guarantee agreement with dataset labels \\
\bottomrule
\end{tabular}
\caption{Systematic divergence patterns between \method entropy-based difficulty estimates and QA evaluation outcomes.}
\label{tbl:qdaps_error_patterns}
\end{table*}

This issue directly affects the entropy of the plausibility scores, and consequently the final difficulty score. For example, as shown in Figure~\ref{fig:gold_answer_inclusion}, the entropy in scenario 1 is $3.3121$, resulting in a normalized difficulty score of $0.99$. However, for the same question and candidate answers in scenario 2, the entropy decreases to $2.5008$, yielding a lower difficulty score of $0.75$.

\newpage

\section{Error Analysis}
\label{apx:error_analysis}

While \method shows strong overall alignment with QA difficulty across datasets, its entropy-based estimates can occasionally diverge from downstream QA evaluation outcomes. This divergence reflects a fundamental distinction between measuring ambiguity in the plausible answer space and measuring success under a specific QA evaluation protocol. In this section, we analyze representative divergence patterns to clarify the conditions under which such misalignments arise.

We refer to these patterns as \emph{entropy-overestimated difficulty} and \emph{entropy-underestimated difficulty}. The former describes cases where \method assigns a high difficulty score (high entropy) despite all QA models answering correctly, while the latter describes cases where \method assigns a low difficulty score (low entropy) despite all QA models failing under exact-match evaluation.

\subsection{Entropy-Overestimated Difficulty}

\paragraph{Question.}
Types of skiing in the Winter Olympics 2018?

\paragraph{Gold Answer.}
Freestyle Skiing.

\paragraph{Analysis.}
For this question (Natural Questions ID: \texttt{nq\_2902}), \method assigns a high difficulty score due to elevated entropy in the candidate plausibility distribution. Multiple skiing-related disciplines receive high plausibility scores, including \emph{Downhill} (0.9442), \emph{Alpine Skiing} (0.9355), \emph{Nordic Combined} (0.925), \emph{Ski Jumping} (0.908), \emph{Cross-Country} (0.901), and \emph{Biathlon} (0.8765), resulting in a broad and competitive plausible answer space.

Despite this high-entropy signal, all evaluated QA models produced correct answers. Model predictions consistently enumerated the appropriate set of Olympic skiing disciplines, differing only in formatting or ordering.

This case exemplifies \emph{entropy-overestimated difficulty}: high entropy reflects the richness of semantically related plausible answers rather than genuine difficulty for QA models. List-based factual questions naturally admit many closely related alternatives, which can inflate entropy even when the task is straightforward for modern QA systems.

\subsection{Entropy-Underestimated Difficulty}

\paragraph{Question.}
What nationality was Aristotle Onassis originally?

\paragraph{Gold Answer.}
Turkish.

\paragraph{Analysis.}
For this question (TriviaQA ID: \texttt{trivia\_10433}), the candidate plausibility distribution is sharply peaked. \emph{Greek} receives a plausibility score of 0.7402, while all other candidates receive substantially lower scores, resulting in low entropy and a low difficulty estimate.

Consistent with this signal, all evaluated QA models unanimously predicted \emph{Greek}. However, the dataset gold answer specifies \emph{Turkish}, reflecting birthplace-based nationality rather than ethnicity or cultural identity. As a result, all model predictions are penalized under evaluation.

This case illustrates \emph{entropy-underestimated difficulty}: \method correctly captures low ambiguity in the plausible answer space, but low entropy does not guarantee agreement with dataset annotations when gold labels encode implicit definitional or historical assumptions.

\subsection{Summary of Divergence Patterns}

Table~\ref{tbl:qdaps_error_patterns} summarizes the two systematic divergence patterns observed between entropy-based difficulty estimation and downstream QA evaluation outcomes.

\subsection{What This Means}

These error patterns clarify the scope and intended interpretation of entropy-based difficulty estimation. \method measures ambiguity in the plausible answer space as perceived by LLMs, rather than the likelihood of producing a response that matches a specific gold annotation under exact-match evaluation. As a result, divergence from QA accuracy is expected in settings with semantically dense answer spaces or annotation-dependent definitions.

Importantly, these cases do not indicate failure of \method; instead, they delineate the boundary between intrinsic question ambiguity and evaluation-specific success criteria. This distinction is particularly relevant for applications such as hallucination risk assessment, question routing, and model selection, where understanding uncertainty in the answer space is more informative than binary correctness.

\newpage

\section{FAQ}\label{apx:faq}

This section addresses common questions readers may have regarding \method and its design choices.

\paragraph{Is this method limited to factoid questions?}
No. As described in Section~\ref{ss:datasets}, \method is evaluated on a diverse set of QA benchmarks. In addition to factoid datasets such as TriviaQA and Natural Questions, we include reasoning-oriented datasets such as QASC and multi-hop reasoning datasets like MuSiQue. The results reported in Section~\ref{ss:model_performance} and Table~\ref{tbl:model_performance} show that \method performs consistently across both simple and complex question types, indicating that it is not restricted to factoid questions.

\paragraph{Does \method heavily rely on the generation quality of the LLM?}
While the quality of the underlying LLM affects candidate generation—as is common in LLM-based approaches—\method does not rely on it exclusively. The ablation study in Table~\ref{tbl:ablation_influence_of_llm} demonstrates that \method maintains strong performance even when using smaller models such as LLaMA~3.1~8B and Qwen~2.5~7B. Across most datasets, \method consistently outperforms baselines regardless of model size, indicating that its effectiveness stems from the plausibility–entropy formulation rather than from a specific high-capacity LLM.

\paragraph{Does the difficulty assessment require significant computational cost?}
No. As analyzed in Section~\ref{ss:answer_plausibility_estimation_methods}, the best-performing configuration adopts the \textit{Listwise} plausibility estimation strategy. This strategy requires only a single prompt per question and therefore has $\mathcal{O}(1)$ prompt complexity. A direct comparison of computational complexity and output length across estimation strategies is provided in Table~\ref{tbl:scenarios_comparison} (Appendix~\ref{apx:answer_plausibility_estimation_methods}), showing that \method is computationally efficient and scalable.

\paragraph{Does the difficulty estimation require access to the correct answer?}
No. \method can operate without access to the gold answer. As shown in the ablation study in Table~\ref{tbl:ablation_without_gold_answers}, removing the gold answer leads to a moderate performance drop but \method still outperforms all baseline methods. This setting reflects realistic scenarios where correct answers are unavailable, confirming that \method does not rely on gold answers to function effectively (see also Appendix~\ref{apx:gold_answer_inclusion} for further analysis).

\paragraph{Does the debiasing step restrict applicability to answers with Wikipedia page views?}
No. Popularity debiasing is an optional refinement rather than a strict requirement. Table~\ref{tbl:ablation_without_debiasing} shows that \method remains competitive and continues to outperform baselines even when the debiasing component is removed. Section~\ref{ss:popularity_correlations} further clarifies that debiasing addresses a specific popularity bias observed during candidate generation. In domains where Wikipedia page views are unavailable or inappropriate, the debiasing step can be safely omitted without compromising the core effectiveness of \method.

\paragraph{What notion of question difficulty does \method capture?}
\method estimates \emph{LLM-oriented difficulty}, defined as the degree of uncertainty exhibited by an LLM when multiple candidate answers appear similarly plausible. This notion is formalized in Section~\ref{s:method} and operationalized via entropy in Section~\ref{ss:score}. Unlike readability-based or retrieval-based difficulty, this definition directly reflects reasoning uncertainty from the model’s perspective.

\paragraph{Why is entropy used instead of average plausibility?}
Entropy captures the distributional spread of plausibility scores rather than collapsing them into a single scalar. As shown in Table~\ref{tbl:model_performance}, entropy-based difficulty estimation consistently outperforms average plausibility across datasets. Section~\ref{ss:score} explains how entropy reflects competition among plausible answers, while Appendix~\ref{apx:case_study} provides a qualitative example motivating this choice.

\paragraph{How stable are difficulty scores across runs and configurations?}
Although candidate generation involves stochasticity, difficulty estimates are stable in aggregate. The robustness of \method across different values of $\alpha$ and different numbers of candidate answers is demonstrated in Appendix~\ref{apx:alpha_robustness} and Appendix~\ref{apx:generalization}. These experiments show consistent performance across a broad range of configurations.

\paragraph{How interpretable are the difficulty scores produced by \method?}
\method is inherently interpretable, as it exposes the candidate answers and their (debiased) plausibility scores that contribute to the final difficulty estimate. A concrete example illustrating this interpretability is provided in Appendix~\ref{apx:case_study}, where each stage of the pipeline and its effect on the final difficulty score is shown step by step.

\newpage
\section{Detailed Results}\label{apx:detailed_results}

This section presents extended results that support and complement the analyses in the main part of the paper. In particular, we provide additional details for Section~\ref{ss:answer_plausibility_estimation_methods} through Table~\ref{tbl:estimation_methods_results_d} (Cohen’s $d$) and Table~\ref{tbl:estimation_methods_results_p} (Spearman’s $\rho$). Similarly, Section~\ref{ss:ablation_study} is supplemented with Table~\ref{tbl:core_results_d} (Cohen’s $d$) and Table~\ref{tbl:core_results_p} (Spearman’s $\rho$).

We also include comprehensive performance breakdowns for a range of baseline approaches, as discussed in Section~\ref{ss:model_performance}. These results are reported in Table~\ref{tbl:model_performance_d} (Cohen’s $d$) and Table~\ref{tbl:model_performance_p} (Spearman’s $\rho$). Each table additionally specifies the corresponding $\alpha$ and sample size $N$ alongside the performance metrics.

\begin{table*}
\centering
\resizebox{\textwidth}{!}{%
\begin{tabular}{l|l|ccc|ccc|ccc|ccc}
\toprule
\textbf{Scenario} & \textbf{Method} & 
\multicolumn{3}{c}{\textbf{MuSiQue}} & 
\multicolumn{3}{c}{\textbf{QASC}} & 
\multicolumn{3}{c}{\textbf{NQ}} & 
\multicolumn{3}{c}{\textbf{TriviaQA}} \\
 &  & $\alpha$ & $N$ & $d$ & $\alpha$ & $N$ & $d$ & $\alpha$ & $N$ & $d$ & $\alpha$ & $N$ & $d$ \\
\midrule
\multirow{2}{*}{Pointwise} 
& Plausibility & -- & 15 & -0.0639 & -- & 6 & \textbf{0.7836} & -- & 13 & 0.9824 & -- & 14 & 0.1786 \\
& Debiased-Plausibility & 0.25 & 15 & \textbf{0.0023} & 0.80 & 14 & 0.6214 & 0.95 & 12 & \textbf{1.098} & 0.82 & 12 & \textbf{0.5557} \\
\midrule
\multirow{2}{*}{Pairwise} 
& Plausibility & -- & 6 & 0.8915 & -- & 6 & 0.1504 & -- & 4 & 0.6016 & -- & 19 & 0.2405 \\
& Debiased-Plausibility & 0.12 & 6 & \textbf{1.1072} & 0.99 & 13 & \textbf{0.3708} & 0.77 & 19 & \textbf{0.7808} & 0.74 & 9 & \textbf{0.3625} \\
\midrule
\multirow{2}{*}{Listwise} 
& Plausibility & -- & 11 & 1.362 & -- & 10 & 0.93 & -- & 15 & 0.9607 & -- & 5 & 0.7775 \\
& Debiased-Plausibility & 0.18 & 9 & \cellcolor{gray!20}\textbf{1.4335} & 0.56 & 14 & \cellcolor{gray!20}\textbf{1.1978} & 0.91 & 20 & \cellcolor{gray!20}\textbf{1.1486} & 0.22 & 4 & \cellcolor{gray!20}\textbf{0.9072} \\
\bottomrule
\end{tabular}
}
\caption{Comparison of plausibility and debiased-plausibility methods across datasets using the Pointwise, Pairwise, and Listwise scenarios. The parameter $\alpha$ indicates the optimal weight used for debiasing popularity, $N$ denotes the optimal number of candidate answers, and $d$ represents the computed Cohen’s $d$ value. \textcolor{gray}{Gray} cells indicate the best value for each dataset, while \textbf{bold} values highlight the best value for each scenario.}
\label{tbl:estimation_methods_results_d}
\end{table*}

\begin{table*}
\centering
\resizebox{\textwidth}{!}{%
\begin{tabular}{l|l|ccc| ccc| ccc| ccc}
\toprule
\textbf{Scenario} & \textbf{Method} & 
\multicolumn{3}{c}{\textbf{MuSiQue}} & 
\multicolumn{3}{c}{\textbf{QASC}} & 
\multicolumn{3}{c}{\textbf{NQ}} & 
\multicolumn{3}{c}{\textbf{TriviaQA}} \\
 & & $\alpha$ & $N$ & $\rho$ & $\alpha$ & $N$ & $\rho$ & $\alpha$ & $N$ & $\rho$ & $\alpha$ & $N$ & $\rho$ \\
\midrule
\multirow{2}{*}{Pointwise} 
& Plausibility & -- & 4 & -0.0545 & -- & 17 & \textbf{-0.5} & -- & 13 & -0.8909 & -- & 17 & -0.0181 \\
& Debiased-Plausibility & 0.39 & 4 & \textbf{-0.1272} & 0.99 & 14 & -0.4848 & 0.10 & 15 & \textbf{-0.9} & 0.98 & 4 & \textbf{-0.309} \\
\midrule
\multirow{2}{*}{Pairwise} 
& Plausibility & -- & 9 & -0.5818 & -- & 4 & -0.4272 & -- & 17 & -0.2818 & -- & 17 & -0.2727 \\
& Debiased-Plausibility & 0.89 & 6 & \textbf{-0.7727} & 0.00 & 4 & \textbf{-0.4272} & 0.99 & 17 & \textbf{-0.5454} & 0.00 & 17 & \textbf{-0.2727} \\
\midrule
\multirow{2}{*}{Listwise} 
& Plausibility & -- & 19 & -0.6454 & -- & 18 & -0.6 & -- & 20 & -0.509 & -- & 19 & -0.5545 \\
& Debiased-Plausibility & 0.93 & 19 & \cellcolor{gray!20}\textbf{-0.8909} & 0.39 & 15 & \cellcolor{gray!20}\textbf{-0.6909} & 0.85 & 19 & \cellcolor{gray!20}\textbf{-0.9636} & 0.75 & 18 & \cellcolor{gray!20}\textbf{-0.6090} \\
\bottomrule
\end{tabular}
}
\caption{Comparison of plausibility and debiased-plausibility methods across datasets using the Pointwise, Pairwise, and Listwise scenarios. The parameter $\alpha$ indicates the debiasing weight, $N$ the number of candidate answers, and $\rho$ the Spearman rank correlation. \textcolor{gray}{Gray} cells indicate the best $\rho$ for each dataset, while \textbf{bold} values highlight the best value within each scenario.}
\label{tbl:estimation_methods_results_p}
\end{table*}

\begin{table*}
\centering
\resizebox{\textwidth}{!}{%
\begin{tabular}{l|l|ccc|ccc|ccc|ccc}
\toprule
\textbf{Model} & \textbf{Method} & 
\multicolumn{3}{c}{\textbf{MuSiQue}} & 
\multicolumn{3}{c}{\textbf{QASC}} & 
\multicolumn{3}{c}{\textbf{NQ}} & 
\multicolumn{3}{c}{\textbf{TriviaQA}} \\
 & & $\alpha$ & $N$ & $d$ & $\alpha$ & $N$ & $d$ & $\alpha$ & $N$ & $d$ & $\alpha$ & $N$ & $d$ \\
\midrule
\multirow{2}{*}{Qwen 2.5 7b} 
 & Plausibility & -- & 5 & 0.6193 & -- & 4 & -0.1797 & -- & 6 & -0.0191 & -- & 7 & 0.1845 \\
 & Debiased-Plausibility & 0.36 & 4 & \textbf{0.8434} & 0.36 & 6 & \textbf{0.1465} & 0.93 & 7 & \textbf{0.2465} & 0.29 & 6 & \textbf{0.3162} \\
\midrule
\multirow{2}{*}{LLaMA 3.1 8b} 
 & Plausibility & -- & 10 & 0.0622 & -- & 19 & 0.144 & -- & 14 & 0.136 & -- & 6 & 0.1754 \\
 & Debiased-Plausibility & 0.51 & 20 & \textbf{0.5467} & 0.92 & 18 & \textbf{0.2484} & 0.01 & 18 & \textbf{0.3886} & 0.04 & 19 & \textbf{0.3481} \\
\midrule
\multirow{2}{*}{LLaMA 3.3 70b} 
 & Plausibility & -- & 9 & 0.894 & -- & 10 & 0.5614 & -- & 6 & 0.88 & -- & 4 & 0.6511 \\
 & Debiased-Plausibility & 0.17 & 6 & \cellcolor{gray!20}\textbf{1.0888} & 0.61 & 12 & \cellcolor{gray!20}\textbf{0.803} & 0.21 & 6 & \cellcolor{gray!20}\textbf{0.9448} & 0.88 & 6 & \cellcolor{gray!20}\textbf{0.7498} \\
\bottomrule
\end{tabular}
}
\caption{Comparison of plausibility vs.\ debiased plausibility methods across datasets for Qwen 2.5 7B, LLaMA 3.1 8b, and LLaMA 3.3 70b used as cores. The parameter $\alpha$ indicates the optimal weight used for debiasing popularity, $N$ denotes the optimal number of candidate answers, and $d$ represents the computed Cohen’s $d$ value. \textcolor{gray}{Gray} cells indicate the best value for each dataset, while \textbf{bold} values highlight the best value for each core.}
\label{tbl:core_results_d}
\end{table*}

\begin{table*}
\centering
\resizebox{\textwidth}{!}{%
\begin{tabular}{l|l|ccc|ccc|ccc|ccc}
\toprule
\textbf{Model} & \textbf{Method} & 
\multicolumn{3}{c}{\textbf{MuSiQue}} & 
\multicolumn{3}{c}{\textbf{QASC}} & 
\multicolumn{3}{c}{\textbf{NQ}} & 
\multicolumn{3}{c}{\textbf{TriviaQA}} \\
 &  & $\alpha$ & $N$ & $\rho$ & $\alpha$ & $N$ & $\rho$ & $\alpha$ & $N$ & $\rho$ & $\alpha$ & $N$ & $\rho$ \\
\midrule
\multirow{2}{*}{Qwen 2.5 7B} 
 & Plausibility & -- & 4 & -0.29 & -- & 20 & 0.3818 & -- & 8 & -0.3181 & -- & 16 & -0.509 \\
 & Debiased-Plausibility & 0.98 & 4 & \textbf{-0.7181} & 0.98 & 11 & \textbf{0.0363} & 0.07 & 6 & \textbf{-0.9636} & 0.28 & 19 & \textbf{-0.7636} \\
\midrule
\multirow{2}{*}{LLaMA 3.1 8B} 
 & Plausibility & -- & 18 & 0.7545 & -- & 5 & -0.4263 & -- & 18 & 0.4545 & -- & 7 & 0.2 \\
 & Debiased-Plausibility & 0.22 & 20 & \textbf{-0.7272} & 0.69 & 19 & \textbf{-0.59} & 0.95 & 19 & \textbf{-0.8454} & 0.41 & 10 & \textbf{-0.8181} \\
\midrule
\multirow{2}{*}{LLaMA 3.3 70B} 
 & Plausibility & -- & 5 & -0.8363 & -- & 15 & -0.4545 & -- & 19 & -0.8909 & -- & 14 & -0.7272 \\
 & Debiased-Plausibility & 0.39 & 6 & \cellcolor{gray!20}\textbf{-0.9001} & 0.54 & 12 & \cellcolor{gray!20}\textbf{-0.6181} & 0.61 & 12 & \cellcolor{gray!20}\textbf{-0.9636} & 0.48 & 12 & \cellcolor{gray!20}\textbf{-0.8818} \\
\bottomrule
\end{tabular}
}
\caption{Comparison of plausibility vs.\ debiased plausibility methods across datasets for Qwen 2.5 7B, LLaMA 3.1 8B, and LLaMA 3.3 70B used as cores. The parameter $\alpha$ indicates the optimal weight used for debiasing popularity, $N$ denotes the optimal number of candidate answers, and $\rho$ the Spearman rank correlation. \textcolor{gray}{Gray} cells highlight the best $\rho$ for each dataset, while \textbf{bold} values mark the best within each core.}
\label{tbl:core_results_p}
\end{table*}

\begin{table*}
\centering
\resizebox{\textwidth}{!}{%
\begin{tabular}{l
  |ccc
  |ccc
  |ccc
  |ccc}
\toprule
\textbf{Category / Method} &
\multicolumn{3}{c}{\textbf{MuSiQue}} &
\multicolumn{3}{c}{\textbf{QASC}} &
\multicolumn{3}{c}{\textbf{NQ}} &
\multicolumn{3}{c}{\textbf{TriviaQA}} \\
 & $\alpha$ & $N$ & $d$
 & $\alpha$ & $N$ & $d$
 & $\alpha$ & $N$ & $d$
 & $\alpha$ & $N$ & $d$ \\
\midrule

\textbf{Readability} \\
\quad Flesch-Kincaid~\cite{rudolf_franz_flesch_1948}
& -- & -- & -0.543
& -- & -- & 0.1496
& -- & -- & -0.424
& -- & -- & -0.2689 \\
\quad Gunning-Fog~\cite{Gunning1952}
& -- & -- & -0.3947
& -- & -- & -0.0944
& -- & -- & -0.5775
& -- & -- & -0.0963 \\
\midrule

\textbf{Prompt-based} \\
\quad LLaMA 3.1 8b~\cite{grattafiori2024llama3herdmodels}
& -- & -- & -0.535
& -- & -- & 0.1065
& -- & -- & 0.0762
& -- & -- & 0.361 \\
\quad LLaMA 3.3 70b~\cite{grattafiori2024llama3herdmodels}
& -- & -- & 0.2453
& -- & -- & 0.2032
& -- & -- & 0.0307
& -- & -- & 0.4566 \\
\midrule

\textbf{Popularity} \\
\quad PopQA~\cite{mallen-etal-2023-trust}
& -- & 20 & -0.0275
& -- & 4 & -0.3206
& -- & 4 & 0.1535
& -- & 16 & -0.2702 \\
\midrule

\textbf{Retriever-based} \\
\quad Retrieval Complexity~\cite{gabburo-etal-2024-measuring}
& -- & -- & 0.1284
& -- & -- & 0.2225
& -- & -- & 0.2781
& -- & -- & 0.4394 \\
\midrule

\textbf{Uncertainty-based} \\
\quad LLaMA 3.1 8b~\cite{2024.EDM-posters.90}
& -- & -- & 0.1365
& -- & -- & 0.1543
& -- & -- & 0.1556
& -- & -- & 0.2211 \\
\quad LLaMA 3.3 70b~\cite{2024.EDM-posters.90}
& -- & -- & 0.4219
& -- & -- & 0.2119
& -- & -- & 0.3265
& -- & -- & 0.4823 \\
\midrule

\textbf{Avg-Plausibility} \\
\quad Plausibility
& -- & 18 & -0.4173
& -- & 10 & 0.2724
& -- & 19 & 0.1663
& -- & 16 & 0.5061 \\
\quad Debiased-Plausibility
& 0.82 & 20 & -0.2242
& 0.53 & 13 & 0.4784
& 0.11 & 19 & 0.1869
& 0.57 & 16 & 0.564 \\
\midrule

\textbf{Entropy-Plausibility} \\
\quad Plausibility
& -- & 9 & \textbf{0.894}
& -- & 10 & \textbf{0.5614}
& -- & 6 & \textbf{0.88}
& -- & 4 & \textbf{0.6511} \\
\quad Debiased-Plausibility
& 0.17 & 6 & \cellcolor{gray!20}\textbf{1.0888}
& 0.61 & 12 & \cellcolor{gray!20}\textbf{0.803}
& 0.21 & 6 & \cellcolor{gray!20}\textbf{0.9448}
& 0.88 & 6 & \cellcolor{gray!20}\textbf{0.7498} \\
\bottomrule
\end{tabular}
}%
\caption{Comparison of various baselines and the \method approach across datasets based on the Cohen's $d$. \textcolor{gray}{Gray} cells indicate the highest score in each dataset, while \textbf{bold} values highlight the second-highest score.}
\label{tbl:model_performance_d}
\end{table*}

\begin{table*}
\centering
\resizebox{\textwidth}{!}{%
\begin{tabular}{l
  |ccc
  |ccc
  |ccc
  |ccc}
\toprule
\textbf{Category / Method} &
\multicolumn{3}{c}{\textbf{MuSiQue}} &
\multicolumn{3}{c}{\textbf{QASC}} &
\multicolumn{3}{c}{\textbf{NQ}} &
\multicolumn{3}{c}{\textbf{TriviaQA}} \\
 & $\alpha$ & $N$ & $\rho$
 & $\alpha$ & $N$ & $\rho$
 & $\alpha$ & $N$ & $\rho$
 & $\alpha$ & $N$ & $\rho$ \\
\midrule

\textbf{Readability} \\
\quad Flesch-Kincaid~\cite{rudolf_franz_flesch_1948}
& -- & -- & 0.5545
& -- & -- & 0.1909
& -- & -- & 0.6363
& -- & -- & 0.5181 \\
\quad Gunning-Fog~\cite{Gunning1952}
& -- & -- & 0.7181
& -- & -- & -0.0636
& -- & -- & 0.6272
& -- & -- & 0.2090 \\
\midrule

\textbf{Prompt-based} \\
\quad LLaMA 3.1 8b~\cite{grattafiori2024llama3herdmodels}
& -- & -- & 0.2636
& -- & -- & -0.1272
& -- & -- & 0.0818
& -- & -- & -0.5545 \\
\quad LLaMA 3.3 70b~\cite{grattafiori2024llama3herdmodels}
& -- & -- & 0.109
& -- & -- & -0.2909
& -- & -- & -0.3363
& -- & -- & -0.4272 \\
\midrule

\textbf{Popularity} \\
\quad PopQA~\cite{mallen-etal-2023-trust}
& -- & 19 & 0.2818
& -- & 4 & 0.2727
& -- & 18 & 0.2636
& -- & 6 & 0.1727 \\
\midrule

\textbf{Retriever-based} \\
\quad Retrieval Complexity~\cite{gabburo-etal-2024-measuring}
& -- & -- & -0.3451
& -- & -- & -0.3126
& -- & -- & -0.4518
& -- & -- & -0.5129 \\
\midrule

\textbf{Uncertainty-based} \\
\quad LLaMA 3.1 8b~\cite{2024.EDM-posters.90}
& -- & -- & -0.3815
& -- & -- & -0.3926
& -- & -- & -0.5025
& -- & -- & -0.3121 \\
\quad LLaMA 3.3 70b~\cite{2024.EDM-posters.90}
& -- & -- & -0.5518
& -- & -- & -0.5621
& -- & -- & -0.5071
& -- & -- & -0.452 \\
\midrule

\textbf{Avg-Plausibility} \\
\quad Plausibility
& -- & 4 & 0.0818
& -- & 14 & 0.0909
& -- & 20 & -0.2545
& -- & 16 & -0.509 \\
\quad Debiased-Plausibility
& 0.02 & 10 & 0.0272
& 0.98 & 10 & -0.3
& 0 & 20 & -0.2545
& 0 & 16 & -0.509 \\
\midrule

\textbf{Entropy-Plausibility} \\
\quad Plausibility
& -- & 5 & \textbf{-0.8363}
& -- & 15 & \textbf{-0.4545}
& -- & 19 & \textbf{-0.8909}
& -- & 14 & \textbf{-0.7272} \\
\quad Debiased-Plausibility
& 0.39 & 6 & \cellcolor{gray!20}\textbf{-0.9001}
& 0.54 & 12 & \cellcolor{gray!20}\textbf{-0.6181}
& 0.61 & 12 & \cellcolor{gray!20}\textbf{-0.9636}
& 0.48 & 12 & \cellcolor{gray!20}\textbf{-0.8818} \\
\bottomrule
\end{tabular}
}%
\caption{Comparison of various baselines and the \method approach across datasets based on the Spearman's $\rho$. \textcolor{gray}{Gray} cells indicate the lowest score in each dataset, while \textbf{bold} values highlight the second-lowest score.}
\label{tbl:model_performance_p}
\end{table*}

\end{document}